\documentclass[preprint,12pt]{elsarticle}

\usepackage{amssymb}
\usepackage{amsmath}

\usepackage{graphicx}
\usepackage{amsmath, amssymb}
\usepackage{array}
\usepackage{booktabs}
\usepackage{hyperref}
\usepackage{float} 
\usepackage{xcolor}
\usepackage{algorithm}
\usepackage{algpseudocode}
\usepackage{subcaption}
\usepackage{adjustbox}
\graphicspath{ {./figures/} }

\journal{Computer Methods in Applied Mechanics and Engineering}

\begin{document}


\begin{frontmatter}



\title{Node Assigned physics-informed neural networks for thermal-hydraulic system simulation: CVH/FL module}


\author[1]{Jeesuk Shin\fnref{eq1}}

\author[2]{Cheolwoong Kim\fnref{eq1}}

\author[3]{Sunwoong Yang}

\author[1]{Minseo Lee}

\author[2]{Sung Joong Kim\corref{cor1}}
\ead{sungjkim@hanyang.ac.kr}

\author[1,4,5]{Joongoo Jeon\corref{cor2}}
\ead{jgjeon41@jbnu.ac.kr}

\cortext[cor1]{Corresponding author}
\cortext[cor2]{Corresponding author}
\fntext[eq1]{These authors contributed equally to this work.}

\affiliation[1]{
  organization={Department of Applied Plasma and Quantum Beam Engineering, Jeonbuk National University},
  city={Jeonju-si},
  country={Republic of Korea}
}

\affiliation[2]{
  organization={Department of Nuclear Engineering, Hanyang University},
  city={Seoul},
  country={Republic of Korea}
}

\affiliation[3]{
  organization={Cho Chun Shik Graduate School of Mobility, Korea Advanced Institute of Science and Technology},
  city={Daejeon},
  country={Republic of Korea}
}

\affiliation[4]{
  organization={Department of Quantum System Engineering, Jeonbuk National University},
  city={Jeonju-si},
  country={Republic of Korea}
}

\affiliation[5]{
  organization={Graduate School of Integrated Energy-AI, Jeonbuk National University},
  city={Jeonju-si},
  country={Republic of Korea}
}

\begin{abstract}
Severe accidents (SAs) in nuclear power plants have been analyzed using thermal-hydraulic (TH) system codes such as MELCOR and MAAP. These codes efficiently simulate the progression of SAs, while they still have inherent limitations due to their inconsistent finite difference schemes. The use of empirical schemes incorporating both implicit and explicit formulations inherently induces unidirectional coupling in multi-physics analyses. The objective of this study is to develop a novel numerical method for TH system codes using physics-informed neural network (PINN). They have shown strength in solving multi-physics due to the innate feature of neural networks---automatic differentiation. We propose a node-assigned PINN (NA-PINN) that is suitable for the control volume approach-based system codes. NA-PINN addresses the issue of spatial governing equation variation by assigning an individual network to each nodalization of the system code, such that spatial information is excluded from both the input and output domains, and each subnetwork learns to approximate a purely temporal solution. In this phase, we evaluated the accuracy of the PINN methods for the hydrodynamic module. In the 6 water tank simulation, PINN and NA-PINN showed maximum absolute errors of 1.678 and 0.007, respectively. It should be noted that only NA-PINN demonstrated acceptable accuracy. To the best of the authors’ knowledge, this is the first study to successfully implement a system code using PINN. Our future work involves extending NA-PINN to a multi-physics solver and developing it in a surrogate manner
\end{abstract}




\end{frontmatter}



\section{INTRODUCTION}\label{sec:1}


Due to the extremely low frequency of severe accident (SA) in nuclear power plants (NPPs) and the limited availability of real-world accident data, SA-related research inevitably relies on the use of system codes to simulate hypothetical accident scenarios and assess the potential safety concerns. Widely used system codes, such as RELAP5/SCDAP, MAAP, and MELCOR, model the physical behavior of NPP components and simulate accident progression by accounting for complex thermal-hydraulic (TH) and physicochemical interactions arising under SA conditions. These simulations cover a broad spectrum of phenomena, including inter-component TH coupling and the release, transport, and deposition of fission products originating from fuel degradation during accident progression~\cite{IAEA_SRS23, IAEA_TECDOC1538}.

Such codes allow users to gain insights into accident phenomena either by analyzing simulation results or validating these results against data acquired from actual accident events~\cite{IAEA_TECDOC1872}. Each system code was developed by a different organization: RELAP5/SCDAPSIM by Innovative Systems Software (ISS) \cite{RELAP5_Manual}, MAAP by the Electric Power Research Institute (EPRI)\cite{MAAP5_Manual}, and MELCOR by Sandia National Laboratories (SNL)\cite{MELCOR_Manual}. For cross-validation purposes, numerous studies have been conducted to compare the outputs of these system codes~\cite{MAAP_MELCOR_Crosswalk1, MAAP_MELCOR_Crosswalk2, CINEMA_MAAP_Comparison}.

Among the available system codes, MELCOR was selected for this study owing to its flexibility in components customization and its suitability for modeling simplified severe accident scenarios ~\cite{MELCOR_Manual}. This distinguished feature makes it a widely used tool for comparative assessments and parametric studies. However, it should be noted that MELCOR exhibits limitations when applied to more complex analyses, such as manual-intensive nodalization, limited multi-physics capabilities and relatively high computational cost.

Recently, deep learning has emerged as a powerful tool for tackling the numerical solution of partial differential equations (PDEs)~\cite{sangam, fvmn}. Among the advances, physics-informed neural network (PINN) have gained particular prominence for capturing the behavior of complex, nonlinear physical systems to this end~\cite{raissi2019physics,yang2024data,mao2020physics}. By embedding the governing physics directly into the loss function in the form of PDEs, PINN differs from conventional black-box deep learning models by offering physically and theoretically interpretable solutions. Notably, as they inherently do not require training data, they can be classified as a numerical method not only machine learning algorithm. Due to their versatility and effectiveness, PINN has garnered considerable attention in a variety of fields, including heat transfer~\cite{heattransferpinn1, heattransferpinn2}, structural~\cite{structuralpinn1, structuralpinn2, structure2} and fluid dynamics~\cite{fluidpinn1, fluidpinn2, fluidpinn3, fluidpinn4, Fluid11, repit}, solid mechanics~\cite{solid1, solid2}, and nuclear reactor kinetics~\cite{nuclear1}, where they have shown promising and efficient alternatives to traditional numerical methods. Once trained, PINN offers notably faster computation compared to traditional numerical methods. However, vanilla PINN still suffer from a key limitation: typically trained for  a single instance, meaning any change in input conditions requires retraining from the scratch. Go et al. \cite{PINN_Surrogate} demonstrated that if trained in the manner of surrogate-PINN, near real-time inference can be utilized within untrained conditions.

\begin{table}[h]
    \centering
    \begin{adjustbox}{max width=\textwidth}
    \begin{tabular}{cccc}
        \toprule
        Publication Year & Author & Used AI & Data-Driven \\
        \midrule
        2021 & Lu et al~\cite{table5}. & ANN & O \\
        2022 & Song et al~\cite{table4}. & LSTM & O \\
        2023 & Chae et al.~\cite{table1} & PINN & O \\
        2023 & Wang et al.~\cite{table2} & ANN & O \\
        2023 & Song et al~\cite{melcorai4}  & LSTM & O \\
        2023 & Antonello et al. ~\cite{Antonello2023}   & PINN & O \\
        2024 & Lee et al~\cite{melcorai2}. & CNN with LSTM & O \\
        2024 & Song et al~\cite{melcorai3}. & CNN with LSTM & O \\
        2025 & Wang et al.~\cite{table3} & ANN & O \\
        2025 & Baraldi et al. ~\cite{Baraldi2025} & PINN & O \\
        \bottomrule
        
    \end{tabular}
    \end{adjustbox}
    \caption{Related Research on AI and Thermal-Hydraulic System Code}
    \label{tab:ai_table}
\end{table}


Although numerous studies have attempted to integrate AI techniques with SA codes, as summarized in Table~\ref{tab:ai_table}, most research has been based on data-driven methodologies and lacks explicit incorporation of physical principles. Lu et al. (2021)~\cite{table5} utilized ANN for predicting the thermal-hydraulic parameters in KLT-40S, a type of nuclear reactor core, and tube-in-tube once-through steam generators. The model developed was trained and validated with RELAP5/SCDAPSIM demonstrating good agreement and accuracy in rapidly predicting thermal-hydraulic parameters. Song et al (2022)~\cite{table4} developed a machine-learning-based simulation model using LSTM to predict severe accident progression at the Fukushima Daiichi Nuclear power plant. The model was trained and validated with MELCOR using observable parameter in main control room such as liquid levels and pressures of reactor core. The study implied that the time series forecasting was appropriately made and the predictions accuracy remained even under the extreme harsh condition such as corrupted measurement device. Chae et al. (2023)~\cite{table1} developed PINN based on the data-driven simulation with the data obtained from MARS-KS, one of the system code developed in South Korea based on RELAP5 code ~\cite{MARS_Manual}. The study suggests the efficiency of the predefined knowledge and the simulation framework could easily be updated. Wang et al. (2023)~\cite{table2} constructed uncertainty analysis tool through bootstrapped ANN by evaluating the severe accidents with the data obtained from MELCOR. Criticizing the heavy load of conventional method of utilizing system code, the bootstrapped ANN demonstrated effective estimation of designated figure of merits such as hydrogen generation and vessel failure timing.
Song et al (2023)~\cite{melcorai4} evaluated the importance of utilizing machine learning especially under the situation of non-available observable parameters. LSTM were utilized in the study. By selecting elimination technique with RandomForestRegressor, the ML model was able to predict the target variable within the true values range. Antonello et al. (2023)~\cite{Antonello2023} analyzed the feasibility of application of PINN with regards to a nuclear battery, a type of unique microreactor, under the case of loss of heat sink accidental scenario. The study compared the two models, DNN and PINN by training the data retrieved from the system code, RELAP-5 3D. The results showed that training the neural network with the domain knowledge provided numerous benefits such as reducing error and avoiding over or underfitting. Lee et al (2024)~\cite{melcorai2} developed a surrogate model for forecasting the progression of the severe accident. The surrogate model was developed based on CNN with LSTM combined. The surrogate model was trained with time series data with thermal-hydraulic behavior. Song et al (2024)~\cite{melcorai3} developed a reinforcement learning with the simulation ran by the surrogate model which was trained with the data obtained from MAAP. The developed reinforcement learning model identified the worst-case scenario for the timing of the high pressure injection failure event occurred. Wang et al. (2025)~\cite{table3} aimed to specifically analyze the quenching process of conical debris beds for Nordic boiling water reactor. The analysis was conducted by coupling the MELCOR system code with an artificial neural network surrogate model. Baraldi et al. (2025)~\cite{Baraldi2025}  developed a physics-informed neural network (PINN)–based surrogate model enhanced with allocation points to accurately estimate key safety parameters during a Loss of Heat Sink (LOHS) scenario in a nuclear microreactor. While the study successfully incorporated multiphysics modeling within a single system, it did not address the multiple control volume conditions. To our best knowledge, this study is the first to introduce a data-free PINN algorithm tailored to MELCOR, developed under multiple control volume and flow path conditions matching general TH analysis.

We propose a novel architecture, called node-assigned PINN (NA-PINN), which naturally aligns with MELCOR’s control volume-based modeling structure. Unlike conventional surrogate models, our approach requires no simulation data, as it embeds the governing equations directly into the loss function. This enables the model to approximate solutions based solely on physical laws, offering a theoretically grounded alternative to traditional numerical solvers. In addition, the resulting predictions are physically interpretable and consistent with fundamental conservation principles. To evaluate the feasibility of the proposed method, we apply it to a simplified but representative scenario involving the MELCOR control volume hydraulic (CVH) / flow path (FL) package, governed by mass and momentum conservation laws. Focusing specifically on the governing physical laws, we assess the convergence characteristics of the PINN architecture within a controlled test environment. The results serve as a foundational step toward constructing physics-informed surrogate models that are compatible with MELCOR-based TH simulations.

The remainder of this paper is organized as follows. Section~\ref{sec:2} describes the emulator for MELCOR's CVH/FL Module, which emulates MELCOR’s thermal-hydraulic calculations, serving as a reference for model validation. Section~\ref{sec:3} introduces the application of PINN to MELCOR simulations, presenting their mathematical formulation and assessing their performance in capturing transient flow behavior. Section~\ref{sec:4} proposes the NA-PINN as an alternative framework to address the limitations of vanilla PINN, detailing its architectural modifications and validating its effectiveness through numerical experiments. Section~\ref{sec:5} discusses potential directions for future research, and Section~\ref{sec:6} concludes the study by summarizing the key findings.







\section{Developing Emulator for MELCOR CVH/FL Module}\label{sec:2}

 \indent A Python-based emulator model is constructed for the evaluation of the comprehension for the calculation in the governing equation of MELCOR, implemented solely using Python code. This emulator facilitates understanding the calculation logic and assessing the accuracy of the governing equations thereby provide the base equation PINN model could train. As a pilot study, the emulator is developed exclusively targeting the CVH and FL modules, intentionally excluding the HS module. Although thermal aspects are typically significant when performing MELCOR calculations, the thermal module was omitted to reduce numerical complexity and clearly demonstrate the feasibility of this methodology before developing it to more complicated scenarios.
 
\begin{figure}[htbp]
    \centering
    \includegraphics[width=0.8\linewidth]{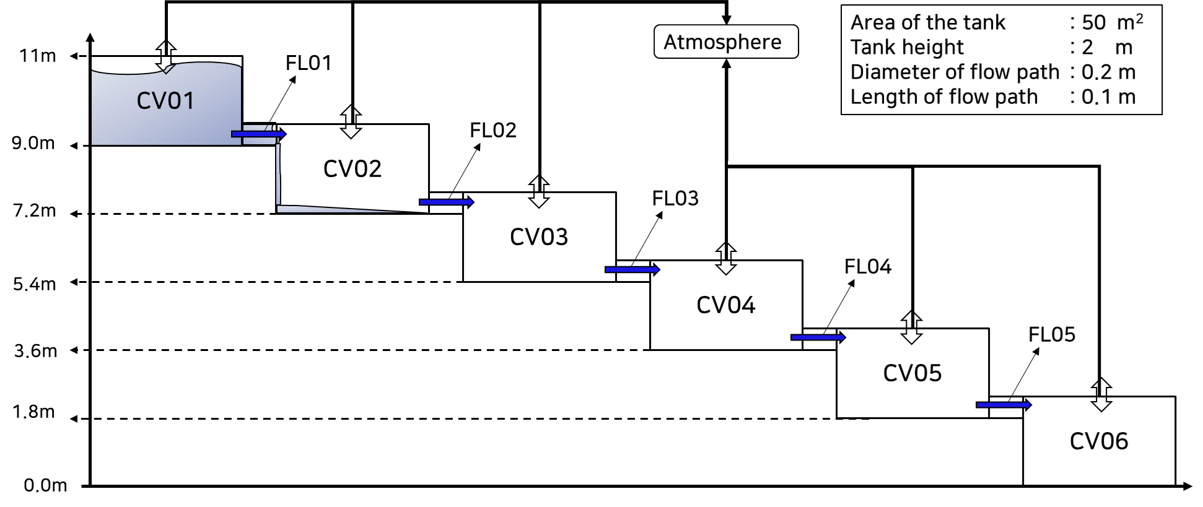}
    \caption{Nodalization for the scenario model}
    \label{fig:example_cwk_1}
\end{figure}

 The selected scenario is demonstrated in Figure~\ref{fig:example_cwk_1}. The water tanks are demonstrated as control volumes (CVs), and the pipe connecting the CVs are demonstrated as FL. In order to neglect the pressure factor within this scenario, all the CVs are assumed to be open on the top implying all the CVs are connected to the atmosphere, respectively. Each tank has an area of 50 m\textsuperscript{2} with a height of 2 m. The diameter of the FL is 0.2 m with length of 0.1 m respectively.

 In this scenario, six CVs are located with different elevation levels of 1.8m, and five FLs are connected at the bottom of the donor tank and top of the receiver tank. In the beginning of the scenario, the highest water tank, CV01 in figure 1, is fully filled with water. As the simulation begins, the water flows toward the bottom of the tank due to gravitational force. The scenario ends when the momentum of the water no longer exists, that is, when the water level of the last two CVs have the same water level. 
 
 \indent To evaluate the accuracy of the comprehension, a Python-based emulator model was developed, hereafter referred to as the \emph{emulator model}. This section provides a detailed explanation of the governing equations implemented in MELCOR and their corresponding computational logic. For clarity, the section is organized into four parts. Section~2.1 introduces the governing equations and defines the associated terminology, focusing on the height and momentum equations used in MELCOR's CVH and FL modules. Section~2.2 discusses the computational logic employed in the emulator. Section~2.3 presents the verification results of the emulator model by comparing the result with that from MELCOR. Finally, Section~2.4 conducts the sensitivity analysis for the equation terms and summarize the key findings with modeling consideration.

\subsection{MELCOR CVH/FL Package: governing equations}\label{sec:2.1}
Two of the equations used in this case scenario are demonstrated in this section. First is the mass conservation equation in partial differential equation form. 

\begin{align}
\frac{\partial M_{i,m}}{\partial t} &= 
\sum_{j} \sigma_{ij} \alpha_{j,\phi} \rho_{j,m}^d v_{j,\phi} F_j A_j + \dot{M}_{i,m}
\label{mass_conservation_1}
\end{align}

The Eq. \eqref{mass_conservation_1}, subscript \(i\) and \(j\) denote the CV of interest and the corresponding FLs respectively. Thus, the summation over subscript \(j\) represents the total mass flow through all associated FLs into or out of the CV\(_i\). The terminology \(\dot{M}_{i,m}\) represent non-flow mass generation processes, including condensation, evaporation, and fog precipitation to name a few. This partial differential equation form can be discretized into the form shown in Eq. \eqref{mass_conservation_2}. Each term in equation.~\eqref{mass_conservation_2} is described in Table \ref{tab:example1}.

\begin{align}
M_{i,m}^n &= M_{i,m}^o 
+ \sum_j \sigma_{ij} \alpha_{j,\phi}^n \rho_{j,m}^d v_{j,\phi}^n F_j A_j \Delta t 
+ \delta M_{i,m}
\label{mass_conservation_2}
\end{align}

\begin{table}[h!]
\centering
\caption{Description of Terms in Equation \eqref{mass_conservation_2}}
\label{tab:example1}
\renewcommand{\arraystretch}{1.5} 
\begin{adjustbox}{max width=\textwidth}
\begin{tabular}{@{}p{1.5cm}p{5.5cm}p{1.5cm}p{5.5cm}@{}}
\toprule
\textbf{Term} & \textbf{Description} & \textbf{Term} & \textbf{Description} \\ \midrule
$M_{i,m}^n$ & New mass of material $m$ in $CV_i$ & $\rho_{j,m}^d$ & Material $m$ density of donor \\
$M_{i,m}^o$ & Old mass of material $m$ in $CV_i$ & $v_{j,\phi}^n$ & Flow velocity calculated from Equation~\eqref{v_equation_form_2} \\
$\sigma_{ij}$ & Direction of flow in FL $j$ & $F_j$ & Fraction of area opened \\
$\alpha_{j,\phi}^n$ & Volume fraction of $\phi$ in FL $j$ & $A_j$ & FL area \\
$\phi$ & Phase of the material (liquid, air, solid) & $\delta M_{i,m}$ & Net external sources \\ 
\bottomrule
\end{tabular}
\end{adjustbox}
\end{table}

 In this scenario, the cross-sectional areas of the CVs and the density of water are assumed to be constant. Therefore, the mass in each CV can be represented in terms of water height. The water height is also used in the momentum equation, specifically to evaluate the static head in the water tank. The partial differential form of the momentum equation is presented in equation~\eqref{v_equation_partial_form_1}. 

\begin{align}
\alpha_{j,\phi} \rho_{j,\phi} L_j \frac{\partial v_{j,\phi}}{\partial t} &=
\alpha_{j,\phi} (P_i - P_k) +
\alpha_{j,\phi} (\rho g \Delta z)_{j,\phi} +
\alpha_{j,\phi} \Delta P_j +
\alpha_{j,\phi} \rho_{j,\phi} v_{j,\phi} (\Delta v_{j,\phi})  \notag \\
&\quad- \frac{1}{2} K_{j,\phi}^* \alpha_{j,\phi} \rho |v_{j,\phi}| v_{j,\phi} 
- \alpha_{j,\phi} \alpha_{j,-\phi} f_{2,j} L_{2,j} (v_{j,\phi} - v_{j,-\phi})
\label{v_equation_partial_form_1}
\end{align}

In equation \eqref{v_equation_partial_form_1}, the subscript \(i\) denotes the donor CV, \(k\) represents the receiver CV, \(\phi\) implies the state of the target material such as liquid or gas, and \(j\) refers to the associated FL. Since MELCOR employs a FDM to solve the momentum equation, the Equation~\eqref{v_equation_partial_form_1} is integrated to derive an approximate discretized form, suitable for numerical implementation. Based on this discretized form, the emulator is developed, and validated by comparing its results with those obtained from MELCOR.

\begin{align}
v_{j,\phi}^n &= v_{j,\phi}^{o+} 
+ \frac{\Delta t}{\rho_{j,\phi} L_j} 
\bigg(
    {P_i^{\tilde{n}}} + \Delta P_j - {P_k^{\tilde{n}}}
    + (\rho g \Delta z)_{j,\phi}^{\tilde{n}} 
    + v_{j,\phi}^{o} \left( \rho \Delta v \right)_{j,\phi}^{o}
\bigg)
\label{v_equation_form_2} \\ \notag
&\quad - \frac{K_{j,\phi} \Delta t}{2 L_j} 
\bigg(
    \big\lvert v_{j,\phi}^{n-} + v_{j,\phi}' \big\rvert v_{j,\phi}^{n} 
    - \big\lvert v_{j,\phi}' \big\rvert v_{j,\phi}^{n-}
\bigg) 
- \frac{\alpha_{j,-\phi} f_{2,j} L_{2,j} \Delta t}{\rho_{j,\phi} L_j} 
\bigg(
    v_{j,\phi}^{n} - v_{j,-\phi}^{n}
\bigg)
\end{align}

In equation~\eqref{v_equation_form_2}, the form- and wall-loss term is derived using a linear approximation based on a Taylor series expansion. Specifically, the loss term in equation~\eqref{v_equation_partial_form_1}, originally expressed as \( |v|v \), is approximated in a linearized form. As a result, the friction loss term in equation~\eqref{v_equation_form_2} is expressed as \( \left| v_{j,\phi}^{n-} + v_{j,\phi}' \right| v_{j,\phi}^{n} - \left| v_{j,\phi}' \right| v_{j,\phi}^{n-} \). 

The choice of \(v_{j,\phi}^{'}\) in this equation depends on the flow direction. If \(v_{j,\phi}^{n-}\) is positive, then \(v_{j,\phi}^{'} = v_{j,\phi}^{n-}\); otherwise, \(v_{j,\phi}^{'}\) is set to be zero. In this scenario, reverse flow is not considered, as no countercurrent flow occurs.

\begin{table}[h!]
    \centering
    \caption{Description of Terms in Equation 4}
    \label{terms_in_v_partial_eqn}
    \renewcommand{\arraystretch}{1.5} 
    \begin{adjustbox}{max width=\textwidth}
    \begin{tabular}{@{}p{1.5cm}p{5.5cm}p{1.5cm}p{5.5cm}@{}}
    \toprule
    \textbf{Term} & \textbf{Description} & \textbf{Term} & \textbf{Description} \\ \midrule
$v_{j,\phi}^n$ & New velocity in $FL_{j}$ & $(\Delta z)_{j,\phi}^n$ & Height of the water \\
$v_{j,\phi}^{o+}$ & Old velocity in $FL_{j}$ & $\Delta P_j$ & Pump head pressure \\
$P_i^{\tilde{n}}$ & Predicted value of pressure of donor volume at end of iteration & $K^{\ast}_{j,\phi}$ & Net form- and wall-loss coefficient \\
$P_k^{\tilde{n}}$ & Predicted value of pressure of receiver volume at end of iteration & $f_{2,j}$ & Momentum exchange coefficient \\
$L_j$ & Inertial length of the pipe & $L_{2,j}$ & Effective length over the interphase force \\
$v_{j,\phi}'$ & Tangent linearization & $v_{j,\phi}^{n-}$ & Old value of $v_{j,\phi}^n$ inside the iteration \\
    \bottomrule
    \end{tabular}
    \end{adjustbox}
\end{table}

\begin{table}[h!]
    \centering
    \caption{Velocity Terms in Momentum Equations}
    \label{terms_in_v_eqn}
    \renewcommand{\arraystretch}{1.5} 
    \begin{adjustbox}{max width=\textwidth}
    \setlength{\tabcolsep}{10pt} 
    \begin{tabular}{@{}clp{8cm}@{}}
    \toprule
    \textbf{No.} & \textbf{Term} & \textbf{Description} \\ \midrule
    1 & $(\rho g \Delta z)_{j,\phi}^{\tilde{n}}$ & Static head difference \\ 
    2 & $v_{j,\phi}^o (\rho \Delta v)_{j,\phi}^o$ & Advection of momentum \\ 
    3 & $\frac{K^*_{j,\phi} \Delta t}{2 L_j} 
        \left( 
        \left| v_{j,\phi}^{n-} + v_{j,\phi}' \right| v_{j,\phi}^n - 
        \left| v_{j,\phi}' \right| v_{j,\phi}^{n-} 
        \right)$ & Net form- and wall-loss effect \\ 
    4 & $\frac{\alpha_{j,-\phi} f_{2,j} L_{2,j} \Delta t}   {\rho_{j,\phi} L_j} 
        \left( v_{j,\phi}^n - v_{j,-\phi}^n \right)$ & Interphase force coefficient (momentum exchange) \\ 
    \bottomrule
    \end{tabular}
    \end{adjustbox}
\end{table}

Table \ref{terms_in_v_partial_eqn} explains the terminologies for equation \eqref{v_equation_partial_form_1} and \eqref{v_equation_form_2}. As per the assumptions, the pressure difference is neglected owing to the open tank. Since there are no force applied in this scenario, the pressure created by pump could also be neglected.
Table \ref{terms_in_v_eqn} demonstrates the effective terms for the velocity calculation. For static head difference term, the elevation difference between the two tanks need to be updated as the time flows during the calculation. The \(\tilde{n}\) indicates the predicted value at the end of calculation step. To be more specific, the static head difference term is explained in the following form:

\begin{align}
{(\rho g \Delta z)}_{j,\phi}^{\tilde{n}} 
= (\rho g \Delta z)_{j,\phi}^{o+} 
+ \frac{\partial (\rho g \Delta z)_{j,\phi}}{\partial M_{i,P}}
\Bigl( M_{i,P}^n - M_{i,P}^{o+} \Bigr)
+ \frac{\partial (\rho g \Delta z)_{j,\phi}}{\partial M_{k,P}}
\Bigl( M_{k,P}^n - M_{k,P}^{o+} \Bigr)
\label{static_head_equation}
\end{align}

Since in this simple case scenario, the amount of loss in donor CV mass is equal to the amount gained from receiver CV mass, the flow can be regarded as consistent and \(\tilde{n}\) can be replaced with \(o\). That is to say, the static head calculation depends only on the value of \(\Delta z\) which is calculated through the height difference between the donor and the receiver. The complexity of such methodology need subtle analysis and assumptions for emulation and is unable to be utilized directly with the equation loss terms for PINN.


\subsection{MELCOR CVH/FL Package: numerical schemes}\label{sec:2.2}
\begin{figure}[H]
    \centering
    \includegraphics[width=1.0\linewidth]{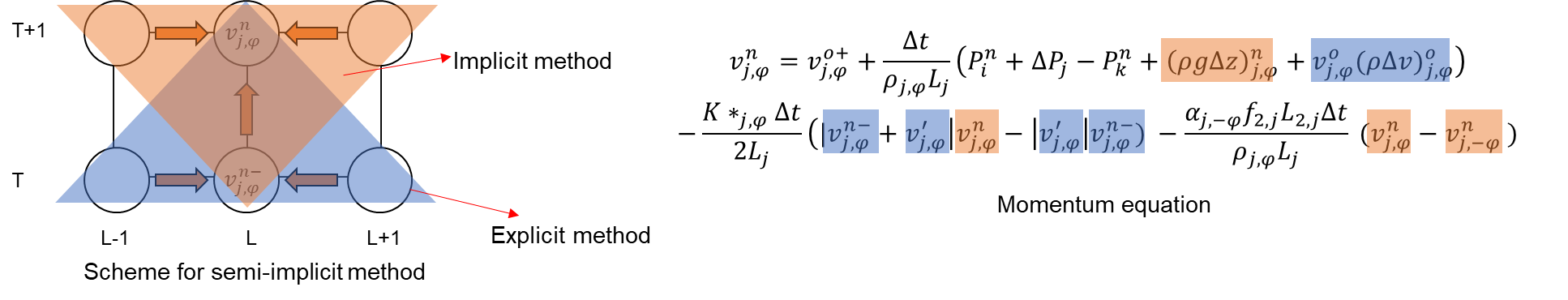}
    
    \caption{Numerical scheme of calculation process utilized in MELCOR}
    \label{fig:numerical_scheme}
\end{figure}

\begin{figure}[H]
    \centering
    \includegraphics[width=0.37\linewidth]{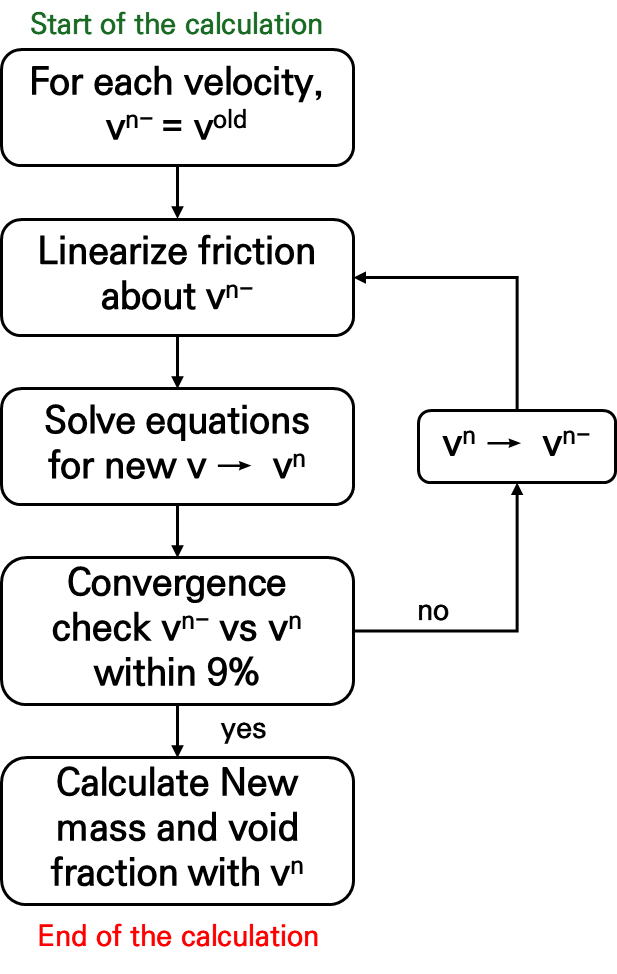}
    
    \caption{Velocity calculation process in MELCOR}
    \label{fig:example_cwk_2}
\end{figure}

Within MELCOR calculation, the terms are retrieved either with implicit method or explicit method. The numerical scheme is demonstrated in Figure~\ref{fig:numerical_scheme}. From the figure, implicit method implies calculating the future value from the future surrounding values whereas explicit method is calculating the future value purely with the current values. From Table \ref{terms_in_v_eqn}, term 1 is calculated implicitly, term 2 is calculated explicitly, term 3 is calculated with both explicitly and implicitly due to mixture of current value and the future value. Term 4 is calculated implicitly. Because the term 3 in Equation \eqref{v_equation_partial_form_1} is nonlinear, the calculation requires iteration until the convergence of the criteria. The calculation process is described in Figure~\ref{fig:example_cwk_2}. The absolute difference between new and old value inside the iteration loop is required to converge within the 9\%.

The calculation begins by adopting the previously computed value, denoted as \(n^-\), and applying a linearization of the friction term around this value.  \(\left| v \right| v\) from equation~\eqref{v_equation_partial_form_1} is approximated linearly with respect to \(n-\). This leads to the expression in equation~\eqref{friction_form_in_v_eqn}:   
 \begin{align} v|v| = v^{n-} |v^{n-}|+(|v^{n-}| ± v^{n-})(v^{n}-v^{n-} ) \label{friction_form_in_v_eqn} \end{align}
where the nonlinear term is expressed in a form suitable for finite difference implementation.

Following the procedures demonstrated in Figure~\ref{fig:example_cwk_2}, the evaluation of the value \(v_{j,\phi}^{n}\) is made with the convergence between the \(v_{j,\phi}^{n-}\)and \(v_{j,\phi}^{n}\) is calculated through iteration. When the convergence is not met, the new value \(v_{j,\phi}^{n}\) is set as ‘\(n^-\)’ and goes into the iteration loop again. Based on this calculation flow, the mass and velocity updates are implemented as described in Algorithm~\ref{alg:mass_update} and ~\ref{alg:velocity_iteration}. The water level calculation is applied to connected tanks $CV_d$ (donor) and $CV_r$ (receiver) with the Algorithm~\ref{alg:mass_update} and the velocity calculation described in Algorithm~\ref{alg:velocity_iteration} applies to each flow path $j = 1,\dots,5$.
MELCOR employs the semi-implicit FDM in CVH/FL packages to ensure the numerical stability and accuracy. Rather than utilizing the global matrix system for the entire system, MELCOR solves the discretized equations locally for coupled FLs, using the iterative methods. Ultimately, the complex form of the iteration and non-iteration forms compose the velocity equation and derives the result of velocity from discretized momentum equation.

\begin{algorithm}[H]
\caption{Mass Update for Coupled Tank System (CVH/FL)}
\label{alg:mass_update}
\begin{algorithmic}[1]
\Require $v^o$, $m_d$, $m_r$, $\Delta t$, $A_p$, $A_t$, $\texttt{elev}$, etc.
\Ensure $m_d^n$, $m_r^n$, $v^n$, $\alpha$

\State Compute donor height:
\[
H_d = \frac{m_d}{\rho A_t} + \texttt{elev}
\]

\If{$H_d - \texttt{elev} \leq \varepsilon$}
    \State $v^n \gets 0$, \quad $\alpha \gets 1.0$
    \State $m_d^n \gets m_d$, \quad $m_r^n \gets m_r$
\Else
    \State Compute the heights to determine the static head:
    \[
    H_d^{\prime} = H_d - \texttt{elev}, \quad H_r = \frac{m_r}{\rho A_t}
    \]
    \If{$H_r < \texttt{elev}$}
        \State ${\Delta z} \gets H_d - \texttt{elev}$ \Comment{Flow into initially empty tank}
    \Else
        \State ${\Delta z} \gets H_d - H_r$ \Comment{Post-alignment inter-tank flow}
    \EndIf

    \State Compute void fraction:
    \[
    \alpha = 
    \begin{cases}
        0, & {H_d} \geq 0.2 \\
        1 - \frac{{H_d}}{0.2}, & 0 \leq {H_d} < 0.2 \\
    \end{cases}
    \]
    \State Compute velocity $v^n$ using Algorithm~\ref{alg:velocity_iteration}
    \State $\Delta m = \rho \cdot v^n \cdot \Delta t \cdot A_p \cdot (1 - \alpha)$
    \State Update: 
    \[
    m_d^n = m_d - \Delta m, \quad m_r^n = m_r + \Delta m
    \]
\EndIf

\Return $m_d^n$, $m_r^n$, $v^n$, $\alpha$, ${\Delta z}$
\end{algorithmic}
\end{algorithm}

\begin{algorithm}[H]
\caption{Velocity Iteration for Given Static Head and Void Fraction}
\label{alg:velocity_iteration}
\begin{algorithmic}[1]
\Require $v^o$, $\alpha$, $\alpha_o$, $\Delta z$, $\Delta t$, $L$, $K$, $A_p$, $A_t$, $\rho$
\Ensure Final velocity $v^n$
\vspace{0.5em}

\State $v^{n-} \gets v^o$, \quad $v_{o2} \gets v^o + (\alpha_o - \alpha)(-v^o)$

\Repeat
    \State $v^{n-} \gets v^{n+}$
    \State Compute:
    \[
    v^{n+} = \frac{
        v_{o2} + \frac{\Delta t}{L} \left( g \Delta z + v_{o2}^2 \cdot \frac{A_p}{A_t} \right)
        + \frac{K \Delta t}{2L} \cdot (v^{n-})^2
    }{
        1 + \frac{K \Delta t}{2L}(v^o + v^{n-}) 
        + \frac{\alpha f L_2 \Delta t}{\rho L} 
        + \frac{\Delta t}{L} \cdot v_{o2} \cdot \frac{A_p}{A_t}
    }
    \]
\Until{$|v^{n+} - v^{n-}| < 0.09 \cdot |v^{n-}|$}

\State \Return $v^n \gets v^{n+}$
\end{algorithmic}
\end{algorithm}

\subsection{Development of Python Emulator model and verification}\label{sec:2.3}

Prior to developing the PINN-based model, it is essential to establish a comprehensive understanding of the governing equations and numerical scheme implemented in MELCOR. Based on the analysis in Sections~2.1 and 2.2, an emulator model was constructed to replicate MELCOR’s behavior and to validate the accuracy of the derived equations. The emulator also offers greater flexibility in manipulating individual terms within the equations, enabling a sensitivity study to identify the dominant contributors. This capability facilitates model simplification by eliminating negligible terms and lays the foundation for efficient training in the subsequent PINN development.

This section presents a comparison between the MELCOR results and those obtained from the developed emulator. the analysis focuses on the two key parameters: FL velocity and water level in the tank. The emulator provides predictions for these parameters, which are compared with MELCOR outputs.

The difference between the two results is quantified using two standard error metrics: mean squared error (MSE) and mean absolute error (MAE). These metrics are defined by the following equations:

\begin{align}
    MAE = \frac{1}{N} \sum_{i=1}^{N} \left| y_i(t) - \hat{y}(t) \right|
    \label{eq:MAE}
\end{align}
\begin{align}
    MSE = \frac{1}{N} \sum_{i=1}^{N} \big( y_i(t) - \hat{y}(t) \big)^{2}
    \label{eq:MSE}
\end{align}

Where N is the number of data, \(y_{i}(t)\) is the desired value (MELCOR values) and \(\hat{y}(t)\) is the predicted value.

 \begin{figure}[H]
    \centering
    \begin{subfigure}{\textwidth}
        \centering
        \begin{subfigure}{0.45\textwidth}
            \centering
            \includegraphics[width=\textwidth]{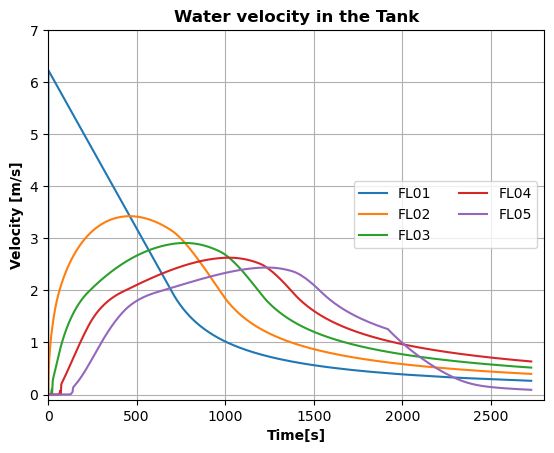}
            \caption{Water level result from MELCOR}
        \end{subfigure}
        \hfill
        \begin{subfigure}{0.45\textwidth}
            \centering
            \includegraphics[width=\textwidth]{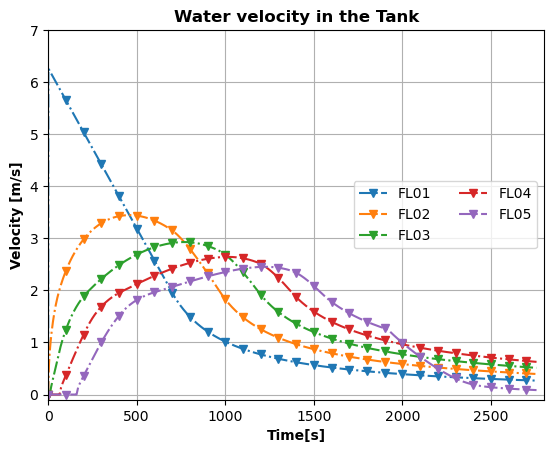}
            \caption{Water level result from Emulator model}
        \end{subfigure}
    \end{subfigure}
    \caption{Velocity profiles in the FL computed by MELCOR (left) and emulator model (right)}
    \label{fig:KCW_velocities}
\end{figure}


A velocity comparison was conducted between MELCOR (target values) and the emulator model (predicted vales). Figure~\ref{fig:KCW_velocities}(a) shows the velocity profile obtained from MELCOR, while Figure~\ref{fig:KCW_velocities}(b) presents the corresponding prediction from the emulator. The overall similarity between the two profiles demonstrates the high accuracy of the emulator.

The computed MAE and MSE over the entire simulation are 0.010412 and 0.00354, respectively. These resulsts indicate that the calculation logic has been successfully reproduced and that the developed emulator is validated for future integration with the PINN framework.

 \begin{figure}[H]
    \centering
    \begin{subfigure}{\textwidth}
        \centering
        \begin{subfigure}{0.45\textwidth}
            \centering
            \includegraphics[width=\textwidth]{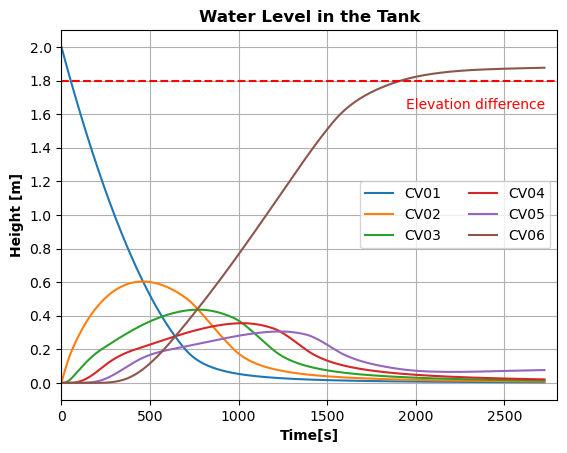}
            \caption{Water level result from MELCOR}
        \end{subfigure}
        \hfill
        \begin{subfigure}{0.45\textwidth}
            \centering
            \includegraphics[width=\textwidth]{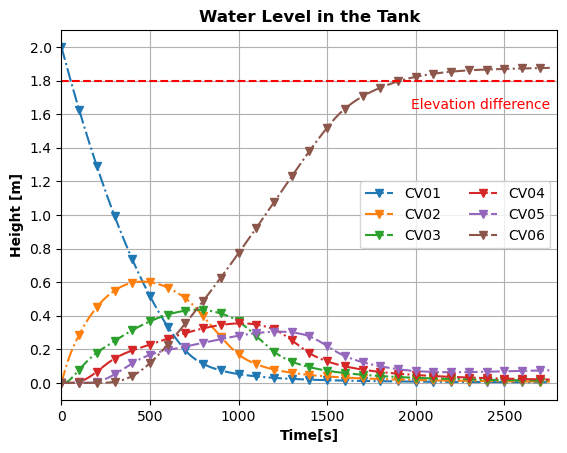}
            \caption{Water level result from Emulator model}
        \end{subfigure}
    \end{subfigure}
    \caption{Water level profiles in the water tanks computed by MELCOR (left) and emulator model (right)}
    \label{fig:KCW_heights}
\end{figure}

Figure~\ref{fig:KCW_heights} depicts the evolution of water level in each tank. Graph (a) presents the reference resuls from MELCOR, while graph (b) shows the predictions from the emulator model. As the water level in the tank decreases, the flow velocity through the FL follows a similar trend. This behavior is governed by Term 1 in Table~\ref{terms_in_v_eqn}, which calculates velocity as a function of the height difference between connected CVs.

A red dotted line is included in the figure to indicate the elevation difference between the two tanks, as defined in Figure~\ref{fig:example_cwk_1}. This elevation marks the point at which the water levels in both tanks become nearly equal, resulting in a sharp reduction in the static head difference and thus a noticeable change in the flow rate. This behavior is reflected in the velocity behavior in FL05, as shown in Figure~\ref{fig:KCW_velocities}.

The emulator's prediction shows a MAE of 0.001571 and a MSE of \(1.1554\times 10^{-5}\) over the entire simulation, indicating reasonable accuracy in terms of mass balance.Table~\ref{tab:error_metrics} presents a comparison of the emulator’s error metrics against MELCOR calculations.

\begin{table}[h!]
\centering
\caption{Comparison of MAE and MSE for velocity and height}
\begin{tabular}{lcc}
\hline
\textbf{Metric} & \textbf{Velocity} & \textbf{Height} \\
\hline
MAE & 0.010412 & 0.001571 \\
MSE & 0.000354 & $1.16 \times 10^{-5}$ \\
\hline
\end{tabular}
\label{tab:error_metrics}
\end{table}

\subsection{Sensitivity Analysis for the equation}\label{sec:2.4}

While MELCOR provides a detailed and validated simulation environment, the modification of individual terms in its governing equations is restricted to the code developers, which limits the possibility of performing term-wise analyses.

One of the key advantages of the emulator is the ability to perform sensitivity analyses by selectively removing or modifying individual terms. In this section, a sensitivity analysis is conducted to identify non-dominant terms that have negligible influence on flow behavior.

This analysis serves as a critical step toward optimizing the governing equation for PINN framework, as it includes the exclusion of unnecessary terms during training, thereby reducing computational cost and improving model reliability.

 \begin{figure}[H]
    \centering
    \begin{subfigure}{\textwidth}
        \centering
        \begin{subfigure}{0.45\textwidth}
            \centering
            \includegraphics[width=\textwidth]{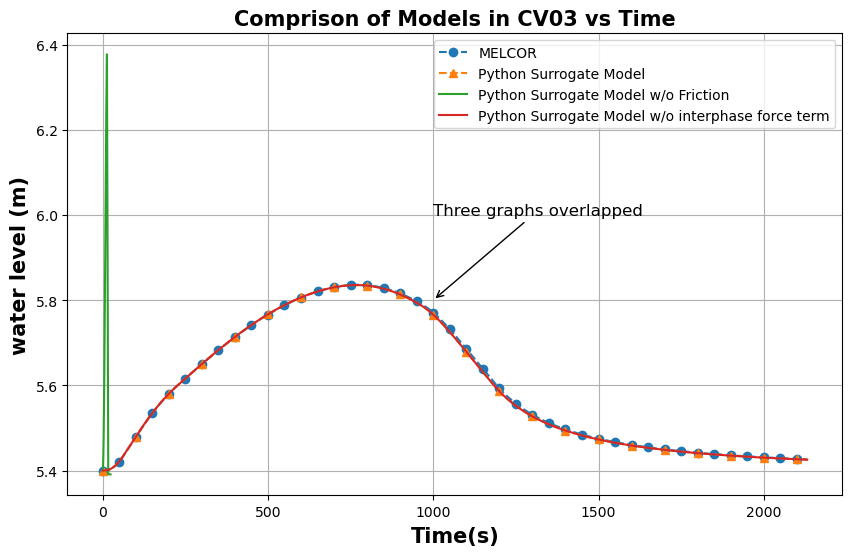}
            \caption{Sensitivity analysis for water level at CV03}
        \end{subfigure}
        \hfill
        \begin{subfigure}{0.45\textwidth}
            \centering
            \includegraphics[width=\textwidth]{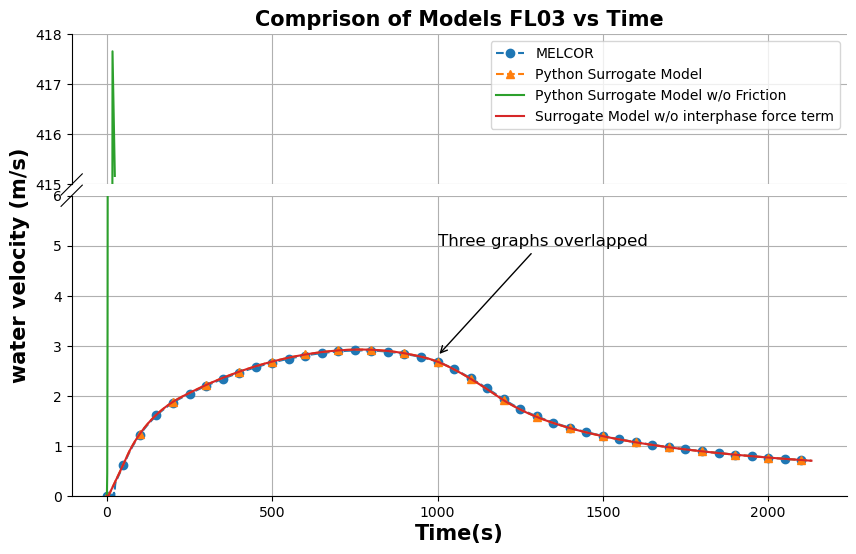}
            \caption{Sensitivity analysis on velocity at FL03}
        \end{subfigure}
    \end{subfigure}
    \caption{Comparison between the models for water level (left) and velocity (right)}
    \label{fig: Sensitivity_analysis}
\end{figure}

As shown in Figure~\ref{fig:KCW_velocities}(b), the emulator demonstrates comparable behavior to MELCOR. To further investigate the governing equation, the velocity-related terms defined in Table~\ref{terms_in_v_eqn} are analyzed, focusing on FL03 and CV03.

Figure~\ref{fig: Sensitivity_analysis}(a) and (b) present the height and velocity responses, respectively, under the selective exclusion of terms 3 and 4 from Table~\ref{terms_in_v_eqn}. Each term was independently removed from the equation to evaluate its influence on the velocity profile.

The analysis reveals that term 3, representing the form or wall loss effect, is the dominant factor influencing the flow. When this term is omitted, the velocity reaches a peak of 417.65~m/s before declining, and the water height rapidly increases to 0.977~m before subsequently decreasing within a short time span. In contrast, excluding term 4 results in no observable difference, indicating that its contribution is negligible for the scenario considered.

This sensitivity analysis allows for the simplification of the model by identifying and eliminating insignificant terms, thereby reducing the computational burden during training. 
Ultimately, the equation that are going to be utilized to train the PINN is the following :

\begin{align}
f_{1} 
= \sum_{j} \alpha_{j,\phi}\, v_{j,\phi}
  \;-\;
  \frac{1}{\rho_{j,m}^d \, A_j} 
  \,\frac{\partial M_{i,m}}{\partial t}
\label{KCW_Mass_conservation}
\end{align}

\begin{align}
f_{2} 
= L_j \,\frac{\partial v}{\partial t}
  \;-\; \bigl(g\,\Delta z\bigr)_{j,\phi}
  \;+\;
  \frac{1}{2}\, K_{j,\phi}^*\,\lvert v\rvert\,v
\label{KCW_Momentum_conservation}
\end{align}

Additionally, a substantial difference in computation time is observed between MELCOR and the emulator. While MELCOR requires 0.5625 seconds to complete the simulation, the emulator model completed the task in 0.03688 seconds. The difference of calculation speed comes from the scope of the calculation. MELCOR contains broader range of tasks such as the heat transfer between the physical models and the behavior of the radionuclides are also included in the scope of MELCOR's work. In contrast, the emulator model has narrowed its focus of calculating the mass and momentum equation, thereby achieving faster computation.

\section{PINN-based CVH/FL Module }\label{sec:3}

\subsection{Background on physics-informed neural network (PINN)}\label{sec:3.1}

PINN represents a pioneering approach that incorporates physical laws directly into neural network architectures by embedding PDEs into the loss function \cite{raissi2019physics}. The fundamental concept of PINN lies in their ability to solve PDEs through neural networks, where the network's loss function includes PDE-based terms that approach zero as the solution converges to satisfy the governing equations. For a well-posed problem, these PDEs are typically complemented by boundary conditions (BCs) and initial conditions (ICs) to ensure solution uniqueness.
PINN can be broadly categorized into two frameworks based on their training approach. The primary focus of this study is on data-free PINN, which solve PDEs without labeled solution data, particularly suitable as alternatives to conventional computational solvers such as MELCOR. In this framework, the neural network is trained solely using the physical constraints imposed by the PDEs and their associated BCs/ICs. While an alternative framework known as data-driven PINN exists, which incorporates additional loss terms from labeled data to enhance prediction accuracy \cite{yang2024data}, our investigation concentrates on data-free PINN due to their potential as standalone computational solvers.
The data-free PINN framework takes spatiotemporal coordinates ($\mathbf{x},t$) as input and predicts the quantities of interest at these points. The physics-based loss function comprises PDE losses and IC/BC losses. For a general PDE of the form:
\begin{equation}
\label{eq:generalPDE}
\mathcal{F}(\mathbf{u}(\mathbf{x},t), \nabla \mathbf{u}(\mathbf{x},t), \nabla^2 \mathbf{u}(\mathbf{x},t), ...) = 0, \quad (\mathbf{x},t) \in \Omega \times [0,T]
\end{equation}
the PDE loss is defined as:
\begin{equation}
\label{eq:PDEloss}
\mathcal{L}_{PDE} = ||\mathcal{F}(\mathbf{u}_{NN}(\mathbf{x},t), \nabla \mathbf{u}_{NN}(\mathbf{x},t), \nabla^2 \mathbf{u}_{NN}(\mathbf{x},t), ...)||,\quad (\mathbf{x},t) \in \Omega \times [0,T]
\end{equation}
where $\mathbf{u}_{NN}$ represents the predicted output values from neural network, $\Omega$ is the spatial domain, and $[0,T]$ is the time interval of interest.
Conventionally, initial and boundary conditions are imposed through additional loss terms. For ICs of the form $\mathbf{u}(\mathbf{x},0) = \mathbf{u}_0(\mathbf{x})$, the IC loss is:
\begin{equation}
\label{eq:ICloss}
\mathcal{L}_{IC} = ||\mathbf{u}_{NN}(\mathbf{x},0) - \mathbf{u}_0(\mathbf{x})||, \quad \mathbf{x} \in \Omega
\end{equation}
Similarly, for BCs of the form $\mathbf{u}(\mathbf{x},t) = \mathbf{g}(\mathbf{x},t)$ on $\partial \Omega$, the BC loss is:
\begin{equation}
\label{eq:BCloss}
\mathcal{L}_{BC} = ||\mathbf{u}_{NN}(\mathbf{x},t) - \mathbf{g}(\mathbf{x},t)||, \quad (\mathbf{x},t) \in \partial\Omega \times [0,T]
\end{equation}
The total PINN loss is then:
\begin{equation}
\label{eq:PINNloss}
\mathcal{L}_{PINN} = \mathcal{L}_{PDE} + \mathcal{L}_{IC} + \mathcal{L}_{BC}
\end{equation}

And there is an alternative way to enforce BCs directly in the network architecture, known as the hard constraint approach \cite{lu2021physics}. In this case, the neural network output $\mathbf{u}_{NN}$ is modified to automatically satisfy the boundary conditions. For BCs of the form $\mathbf{u}(\mathbf{x},t) = \mathbf{g}(\mathbf{x},t)$ on $\partial \Omega$, the modified output is:
\begin{equation}
\label{eq:hardBC}
\mathbf{u}(\mathbf{x},t) = h(\mathbf{x}) \cdot \mathbf{u}_{NN}(\mathbf{x},t) + \mathbf{g}(\mathbf{x},t)
\end{equation}
where $h(\mathbf{x})$ is a function that equals 0 on $\partial\Omega$. With the hard constraint approach for BCs/ICs, the PINN's loss only includes the PDE term since they already satisfy BCs/ICs:
\begin{equation}
\label{eq:hardPINNloss}
\mathcal{L}_{PINN} = \mathcal{L}_{PDE}
\end{equation}

In contrast to the multiplicative formulation presented in Eq~\eqref{eq:hardBC}, this study adopts an additive strategy, hereafter referred to as the "shifting" method. Further details on this formulation are provided in Section~\ref{sec:3.2} 

All partial derivatives in the equations for the loss functions of the PINN are computed using automatic differentiation within the neural network \cite{raissi2019physics}. The loss terms are evaluated at spatiotemporal collocation points: PDE losses within the domain $\Omega \times [0,T]$ and IC/BC losses at their respective locations. Naturally, the strategic selection of these collocation points is known to significantly influence the PINN's performance \cite{lu2021deepxde,nabian2021efficient,mao2020physics, yang2024data}. After training, PINN provide significantly faster computational performance than conventional numerical methods. Although vanilla PINN is typically limited to single-instance applications, this speed advantage can be leveraged when they are trained in the style of surrogate PINN.



\subsection{Reformulation of PINN for MELCOR}\label{sec:3.2}

The vanilla PINN architecture is formulated as a fully connected neural network (FCNN), where the input variable is the one-dimensional time variable $t$. Unlike general PDE solvers such as CFD, which require spatial coordinates and compute continuous fields of solution variables, MELCOR calculates solution variables individually for each CV due to the nature of the system code. As a result, spatial coordinates are not needed in this context, and the network must predict the solution for each CV separately, rather than as a continuous spatial field. The output dimension of the network is therefore defined as $2N_i - 1$, where $N_i$ denotes the number of CVs. Specifically, the network simultaneously predicts $N_i$ variables representing the water heights of the CVs, along with $N_i - 1$ variables corresponding to the flow velocities between adjacent CVs: see Figure~\ref{fig:example_cwk_1}.

\begin{figure}[h]
    \centering
    \includegraphics[width=1.0\textwidth]{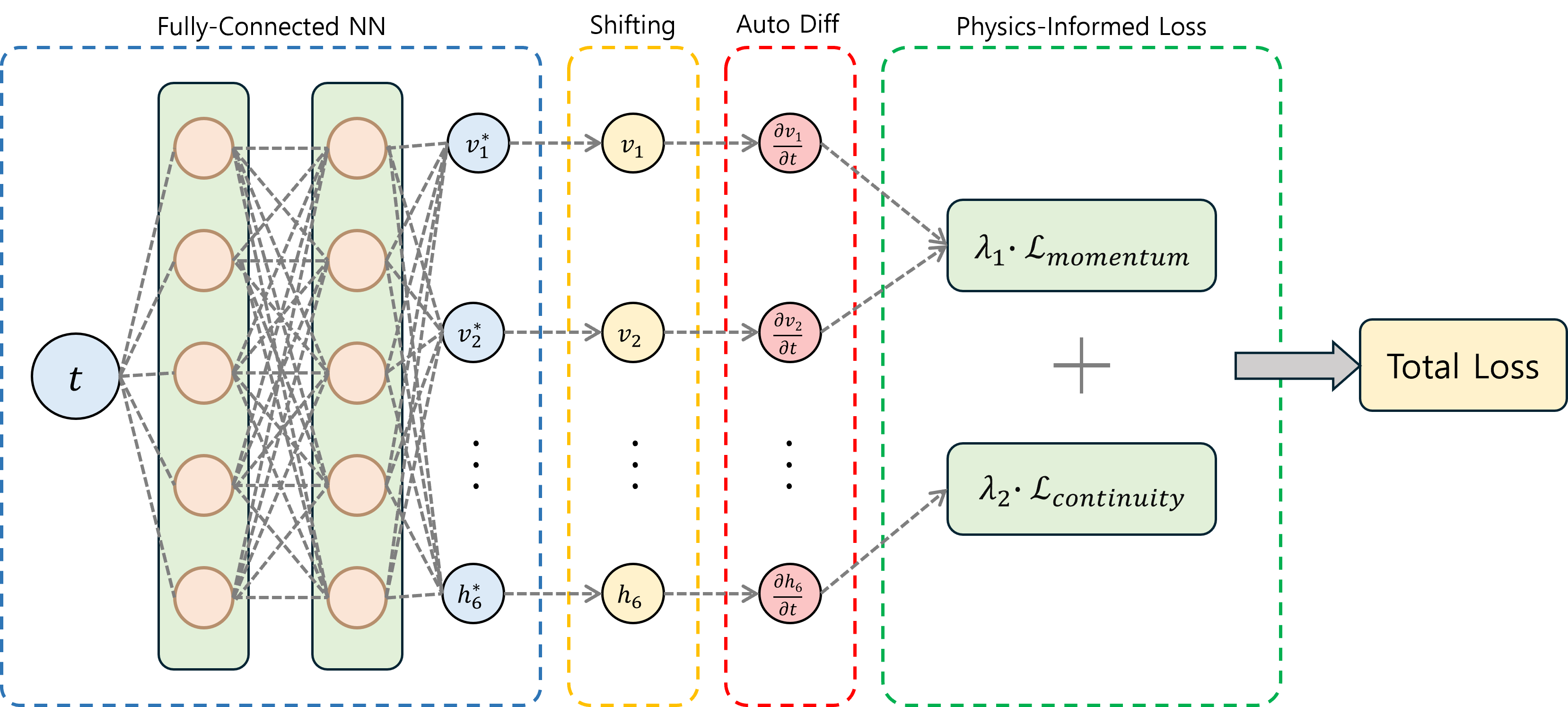}
    \caption{Architecture of the vanilla PINN model}
    \label{fig:pinn_architecture}
\end{figure}

This integrated network structure is essential due to the coupling between the CVs and the FLs. The water height in each CV depends on the velocities of the connected FLs, whereas the velocity in each FL is governed by the water height difference between adjacent CVs. Due to these mutual dependencies, the network outputs for water heights and velocities must be trained simultaneously to properly capture their coupled behavior.Therefore, a single network is used to simultaneously learn the fully coupled dynamics of the system. The overall PINN framework used in this study is illustrated in Fig.~\ref{fig:pinn_architecture}.

To strictly enforce the initial condition, while no boundary conditions are considered in this study, a shifting method based on an additive formulation was adopted.As shown in Eq~\eqref{eq:hard_constraints}, the network output is reformulated by adding a correction term to ensure that the predefined initial condition $\mathbf{u}_0$ is exactly satisfied. This additive approach is preferred over the conventional multiplicative formulation, as it imposes the initial condition without introducing additional functional complexity to the network output. In this formulation, $\mathbf{u}_{NN}(t)$ denotes the original network output at time $t$, $\mathbf{u}_{NN}(0)$ represents the network prediction at the initial time 
$t = 0$, and $\mathbf{u}_0$ is the given initial condition. By directly incorporating the initial condition into the network architecture, it is inherently satisfied throughout the training process.

\begin{equation}
    \mathbf{u}(t) = \mathbf{u}_{NN}(t) - \mathbf{u}_{NN}(0) + \mathbf{u}_0.
    \label{eq:hard_constraints}
\end{equation}

Since the initial condition is strictly satisfied through the hard constraint formulation and no boundary conditions are imposed, the total loss function $\mathcal{L}_{PDE}$ is defined solely by the momentum conservation loss $\mathcal{L}_{momentum}$ and the continuity equation loss $\mathcal{L}_{continuity}$. This is expressed in Eq~\eqref{eq:total_lossfunction} as

\begin{align}
    \mathcal{L}_{PDE} = \lambda_1 \cdot \mathcal{L}_{momentum} + \lambda_2 \cdot \mathcal{L}_{continuity}
    \label{eq:total_lossfunction}
\end{align}
where the weighting parameters $\lambda_1 = 1$ and $\lambda_2 = 0.001$ were empirically selected to balance the contributions of each term and enhance numerical stability.

The momentum conservation loss $\mathcal{L}_{momentum}$, derived from the governing momentum equation, is calculated as shown in Eq~\eqref{eq:loss_momentum}. In this expression, $N_j$ denotes the number of pipes, defined as $N_i - 1$, and $v_{j,\phi}$ represents the velocity of the $j$-th FL. As noted above, CV-based approach was used in MELCOR code. In addition, ${N_t}$ refers to the number of collocation points at which the residuals of both the momentum and continuity equations are evaluated during training.

\begin{align}
    \mathcal{L}_{momentum} &= \frac{1}{N_t} \sum_{j=1}^{N_j} \left(L\frac{\partial v_{j,\phi}}{\partial t} - g \Delta z + \frac{1}{2} K_{j,\phi}^* |v_{j,\phi}| v_{j,\phi} \right)^2
    \label{eq:loss_momentum}
\end{align}

For mass conservation, the continuity equation is reformulated under the incompressible flow assumption. The mass $M_{i,m}$ within the $i$-th CV is expressed in Eq~\eqref{eq:transform_m_to_h}, where $H_{i,m}$ represents the water height in the CV, $A_i$ is the cross-sectional area of the CV base, and $\rho_{i,m}^d$ denotes the fluid density. Using this formulation, the continuity residual for each CV is defined by Eq~\eqref{eq:new_continuity_pde}, where $\sigma_{ij}$ indicates the flow direction, $\alpha_{j,\phi}$ is the flow area ratio, $F_j$ is the flow coefficient, and $A_j$ is the cross-sectional area of the FL.

\begin{align}
    M_{i,m} &= H_{i,m} A_i \rho_{i,m}^d
    \label{eq:transform_m_to_h}
\end{align}

\begin{align}
    A_i \rho_{i,m}^d \frac{\partial H_{i,m}}{\partial t} &= 
    \sum_{j} \sigma_{ij} \alpha_{j,\phi} \rho_{j,m}^d v_{j,\phi} F_j A_j
    \label{eq:new_continuity_pde}
\end{align}

To account for different flow characteristics depending on the CV location, the continuity residual is classified into three cases:\par
1. In the first CV, water can only exit through the connected pipe, leading to Eq.~\eqref{eq:first_cv_contiunity}.\par
2. In the last CV, water can only enter through the connected pipe, satisfying Eq.~\eqref{eq:last_cv_contiunity}.\par
3. For an intermediate CV (neither the first nor the last), water flows in from the preceding CV and exits to the next CV, which results in Eq.~\eqref{eq:middle_cv_contiunity}.

\begin{align}
    \mathcal{R}_{1} &=  A_i \rho_{i,m}^d \frac{\partial H_{i,m}}{\partial t} + \rho_{j,m}^d v_{j,\phi} F_j A_j 
    \label{eq:first_cv_contiunity}
\end{align}

\begin{align}
    \mathcal{R}_{i} &=  A_i \rho_{i,m}^d \frac{\partial H_{i,m}}{\partial t} - \rho_{j-1,m}^d v_{j-1,\phi} F_{j-1} A_{j-1}
    \label{eq:last_cv_contiunity}
\end{align}

\begin{align}
    \mathcal{R}_{2} \sim \mathcal{R}_{i-1} &=  A_i \rho_{i,m}^d \frac{\partial H_{i,m}}{\partial t} + \rho_{j,m}^d v_{j,\phi} F_j A_j - \rho_{j-1,m}^d v_{j-1,\phi} F_{j-1} A_{j-1} 
    \label{eq:middle_cv_contiunity}
\end{align}

The total continuity loss $\mathcal{L}_{continuity}$ is calculated as the mean squared sum of the local continuity residuals across all CVs, where each residual $\{ \mathcal{R}_i \}_{i=1}^{N_i}$ is defined according to the flow characteristics of the first, last, and intermediate CVs in Eqs.~\eqref{eq:first_cv_contiunity}--\eqref{eq:middle_cv_contiunity}. The final form of the continuity loss is given in Eq~\eqref{eq:loss_contiunity}.

\begin{align}
    \mathcal{L}_{continuity} &= \frac{1}{N_t} \sum_{i=1}^{N_i} \mathcal{R}_{i}^2
    \label{eq:loss_contiunity}
\end{align}

For network training, the ELU activation function \cite{clevert2015fast} was used to prevent vanishing gradient issues, and He initialization was applied for the initialization of the weight. The learning rate was set to 0.0001, with a scheduler reducing the learning rate by 99.99\% at each epoch. In addition, Min-max scaling is applied to the time variable $t$ to maintain numerical stability during training by normalizing the input range to [0, 1], as shown in Eq~\eqref{eq:min_max_scaling}, where $t_{\text{min}}$ and $t_{\text{max}}$ denote the minimum and maximum time values of the collocation points.

\begin{equation}
    t' = \frac{t - t_{\min}}{t_{\max} - t_{\min}},
    \label{eq:min_max_scaling}
\end{equation}

The complete training algorithm is provided in Algorithm~\ref{al:pinn_algorithm}. Since the material $m$ and phase $\varphi$ do not change in this scenario, they are excluded for simplicity.

\begin{algorithm}
\scriptsize
\caption{Training procedure of PINN}\label{al:pinn_algorithm}
\begin{algorithmic}[1]
\setlength{\itemsep}{-1.5pt}
\Require Time domain \( [0, T] \), number of collocation points \( N_t \), initial condition \( \mathbf{u}_0 \)

\Require Trainable parameters \( \boldsymbol{\theta} \), number of tanks \( n \), number of epochs \( N_{\text{epoch}} \), learning rate scheduler \( \eta_{\text{sched}} \)
\State Initialize neural network parameters \( \boldsymbol{\theta} \)
\State Apply shifting method:
\[
    \mathbf{u}(t) = \mathbf{u}_{NN}(t) - \mathbf{u}_{NN}(0) + \mathbf{u}_0
\]
\For{each epoch \( k = 1, \dots, N_{\text{epoch}} \)}
    \State Sample \( N_t \) collocation points \( t \in [0, T] \) using uniform (equally spaced) discretization
    \State Normalize time input using Min-Max scaling:
    \[
        t' = \frac{t - t_{\min}}{t_{\max} - t_{\min}}
    \]
    \State Predict \( \{ \hat{v}_{j}(t') \}_{j=1}^{n-1} \) and \( \{ \hat{H}_{i}(t') \}_{i=1}^{n} \) using the neural network $\mathbf{u}(t)$
    \State Compute temporal derivatives via automatic differentiation:
    \[
        \frac{\partial \hat{v}_{j}}{\partial t},\quad \frac{\partial \hat{H}_{i}}{\partial t}
    \]
    \State Compute void fraction at the inlet of pipe \( j \) (with tank index \( i = j \)):
    \[
    \alpha_j =
    \begin{cases}
        1 - \dfrac{\hat{H}_{j}}{0.2}, & 0 \leq \hat{H}_{j} < 0.2 \\
        0, & \text{otherwise}
    \end{cases}
    \]
    \State Compute height difference for each pipe \( j \) (with tank index \( i = j \)):
    \[
    \Delta z_j =
    \begin{cases}
        \hat{H}_{j} - \hat{H}_{j+1} + 1.8, & \hat{H}_{j} < 0.2 \land \hat{H}_{j+1} \ge 1.8 \\
        \hat{H}_{j}, & \text{otherwise}
    \end{cases}
    \]
    \State Compute momentum residuals for each pipe \( j \):
    \[
        \mathcal{R}_{\text{momentum}, j} = L \cdot \frac{\partial \hat{v}_{j}}{\partial t} - g \cdot \Delta z_j + \frac{K^*}{2} |\hat{v}_{j}| \hat{v}_{j}
    \]
    \State Compute continuity residuals for each tank \( i \) (with pipe index \( j = i \)):
    \[
    \mathcal{R}_{\text{continuity}, i} = \rho A_t \cdot \frac{\partial \hat{H}_{i}}{\partial t} +
    \begin{cases}
    \rho A_p F \cdot \hat{v}_{1} (1 - \alpha_1), & i = 1 \\
    \rho A_p F \cdot \left[ \hat{v}_{i} (1 - \alpha_i) - \hat{v}_{i-1} (1 - \alpha_{i-1}) \right], & 2 \le i \le n{-}1 \\
    -\rho A_p F \cdot \hat{v}_{n-1} (1 - \alpha_{n-1}), & i = n
    \end{cases}
    \]
    \State Compute total loss:
    \[
        \mathcal{L}_{\text{PDE}} = \lambda_1 \cdot \frac{1} {N_t}\sum_{j=1}^{n-1} \mathcal{R}_{\text{momentum}, j}^2 + \lambda_2 \cdot \frac{1}{N_t}\sum_{i=1}^{n} \mathcal{R}_{\text{continuity}, i}^2
    \]
    \State Update parameters using a scheduled learning rate:
    \[
        \boldsymbol{\theta} \leftarrow \boldsymbol{\theta} - \eta_{\text{sched}} \nabla_{\boldsymbol{\theta}} \mathcal{L}_{\text{PDE}}
    \]
\EndFor
\State \Return trained model \( \mathbf{u}(t) \)
\end{algorithmic}
\end{algorithm}

Through this formulation, the proposed PINN architecture is designed to capture the coupled dynamics of CVs and FLs while strictly adhering to the governing physical laws. However, the effectiveness of this approach must be verified across various system configurations to assess its accuracy and scalability. Therefore, in the following section, a series of case studies are conducted under different CV settings to evaluate the performance of the proposed model.

\subsection{Case study matrix}\label{sec:3.3}

In this section, the performance and scalability of the vanilla PINN architecture are evaluated through a series of numerical experiments. The objective of these case study is to assess the model’s ability to generalize across systems with varying complexity due to differing numbers of CVs. To this end, simulations are conducted for systems comprising $N_i = 2$, $N_i = 3$, and $N_i = 6$, where $N_i$ denotes number of CVs.

All case studies are conducted under identical physical conditions as defined in Section~\ref{sec:2}. However, the termination criteria differ due to the fundamental differences in how the models handle temporal evolution. The Python-based emulator model follows a step-by-step computation from the initial condition, where the simulation progresses sequentially in time until a physical convergence condition is met—for example, when the water heights of the final CV and its preceding CV become sufficiently close.

In contrast, the PINN model learns the solution over the entire time domain simultaneously, rather than advancing in discrete time steps. As a result, it requires an explicitly predefined termination time to define the temporal learning domain. Furthermore, as the number of CVs increases, the scenario duration naturally becomes longer, and the termination time for PINN must be extended accordingly.

To ensure consistency between the models, the termination time for each PINN case was set by rounding up the final simulation time of the emulator model to the nearest hundred. For example, in the case with six CVs, the emulator's simulation terminated at 2727 seconds, so the corresponding termination time for PINN was set to 2800 seconds.

To account for the increased system complexity associated with larger numbers of CVs, the network architecture and training parameters are scaled accordingly. Specifically, the number of collocation points, training epochs, and the size of the neural network increase proportionally with $N_i$. Collocation points are uniformly placed at intervals of approximately 0.4 seconds to ensure sufficient temporal resolution. The detailed hyperparameter settings for each configuration are summarized in Table~\ref{tab:table_of_hyperparameters_of_pinn}, with corresponding results discussed in the subsequent section.

\begin{table}[h!]
    \centering
    \begin{adjustbox}{max width=\textwidth}
    \begin{tabular}{ccccccc}
        \toprule
        \textbf{$N_i$} & \textbf{End Time} & \textbf{Collocation Points} & \textbf{Epochs} & \textbf{Hidden Layers $\times$ Nodes} \\
        \midrule
        2 & 1000 & 2500 & 30000 & $10 \times 192$ \\
        3 & 1400 & 3000 & 40000 & $10 \times 256$ \\
        6 & 2800 & 6000 & 50000 & $10 \times 368$ \\
        \bottomrule
    \end{tabular}
    \end{adjustbox}    
    \caption{Hyperparameters for each case according to the number of CVs.}
    \label{tab:table_of_hyperparameters_of_pinn}
\end{table}
\subsection{PINN module verification}\label{sec:3.4}

A comparative analysis was conducted to evaluate the accuracy of the proposed PINN by examining its predictions against those of the emulator model. The evaluation considers water height and velocity profiles across three different CV configurations ($N_i = 2$, $3$, and $6$), allowing for a systematic investigation of the model's limitations under increasing system complexity.

The accuracy of the PINN model is quantitatively assessed using the mean absolute error (MAE) and mean squared error (MSE), calculated according to Equations~(\ref{eq:MAE}) and (\ref{eq:MSE}), respectively. These metrics provide a quantitative measure of the discrepancies between the PINN predictions and the reference solutions.

\begin{figure}[H]
    \centering
    \begin{subfigure}{\textwidth}
        \centering
        \begin{subfigure}{0.45\textwidth}
            \centering
            \includegraphics[width=\textwidth]{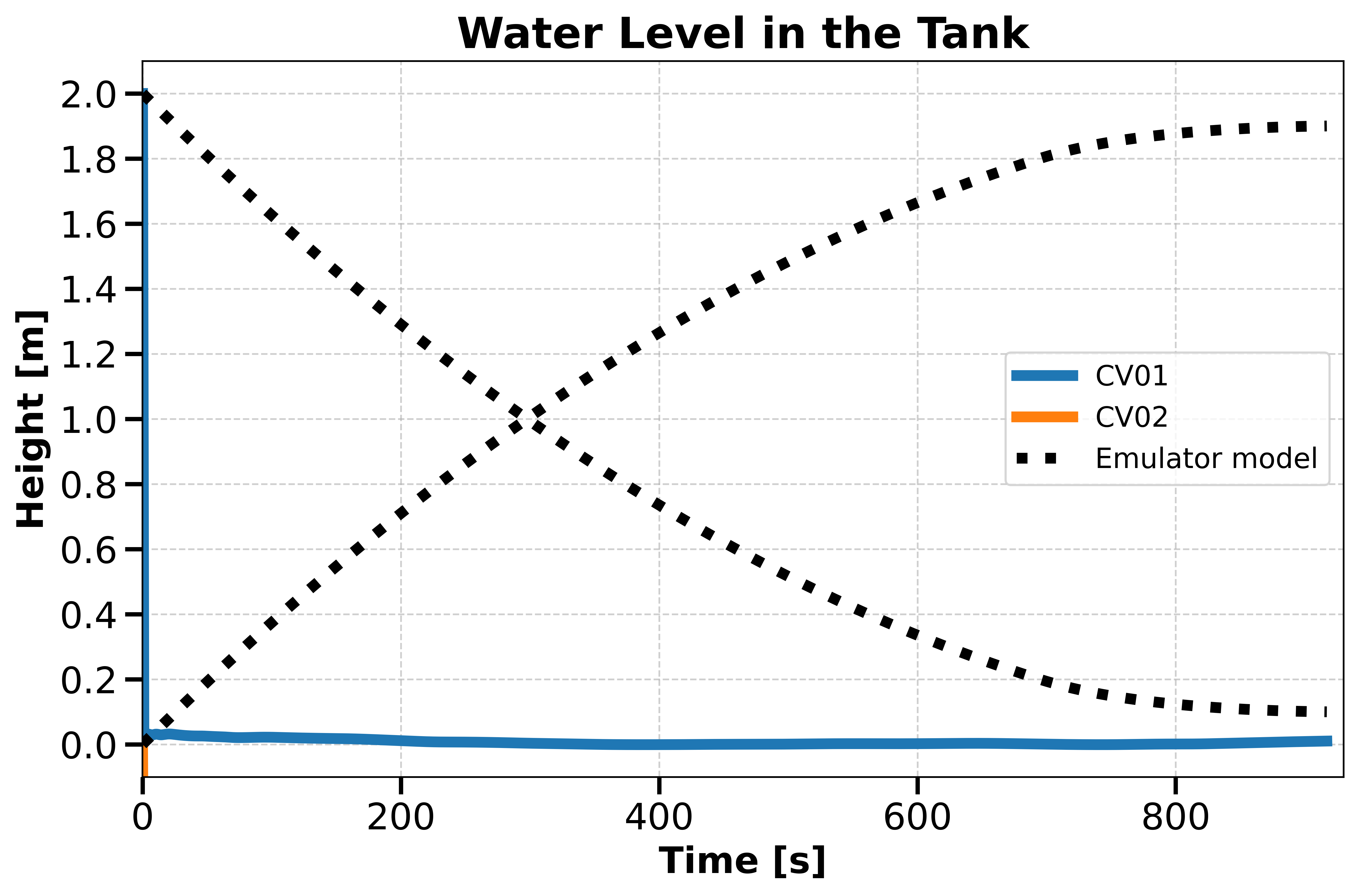}
            \caption{Height profile when $N_i=2$}
        \end{subfigure}
        \hfill
        \begin{subfigure}{0.45\textwidth}
            \centering
            \includegraphics[width=\textwidth]{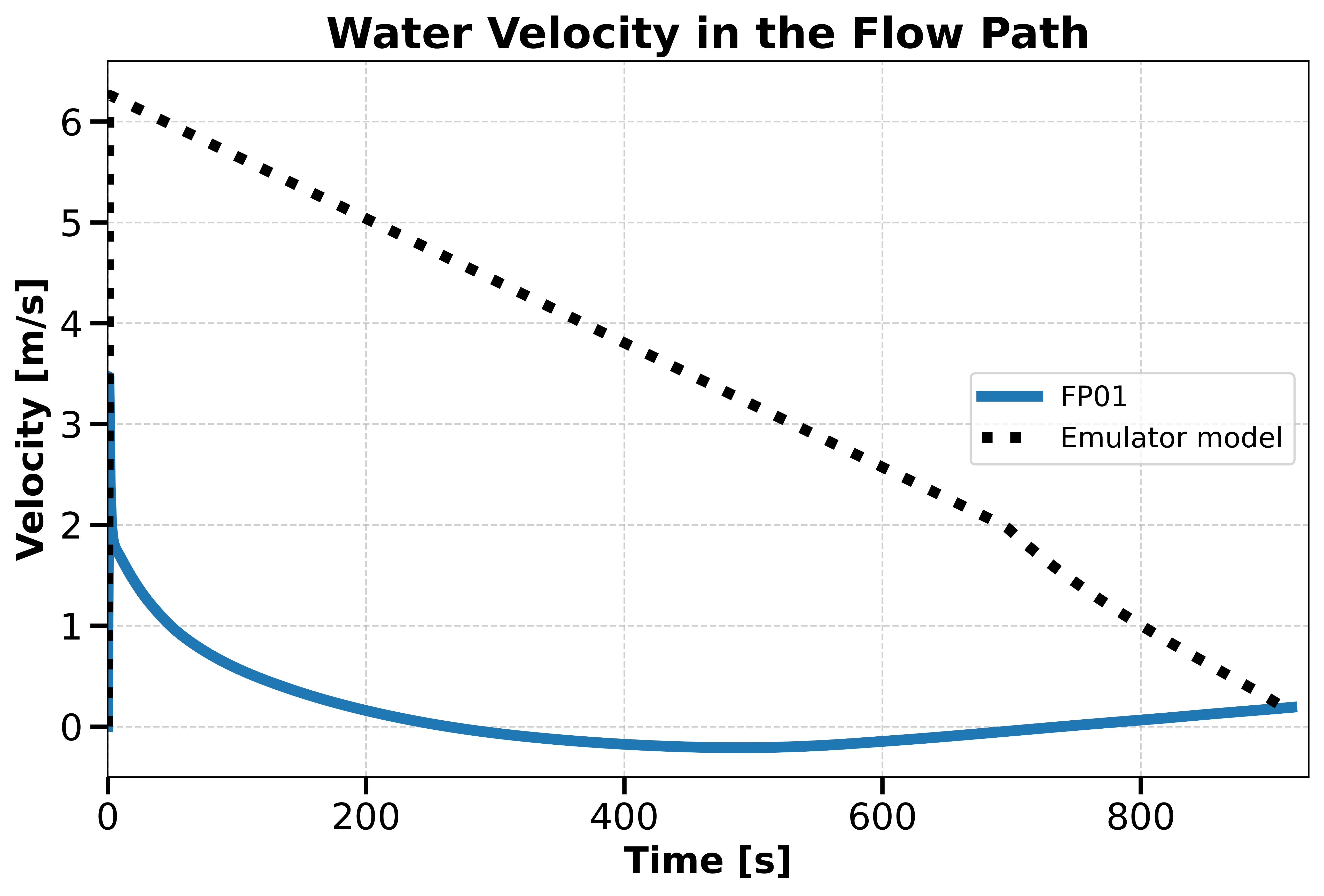}
            \caption{Velocity profile when $N_i=2$}
        \end{subfigure}
    \end{subfigure}
    
    \vspace{0.3cm}
    
    \begin{subfigure}{\textwidth}
        \centering
        \begin{subfigure}{0.45\textwidth}
            \centering
            \includegraphics[width=\textwidth]{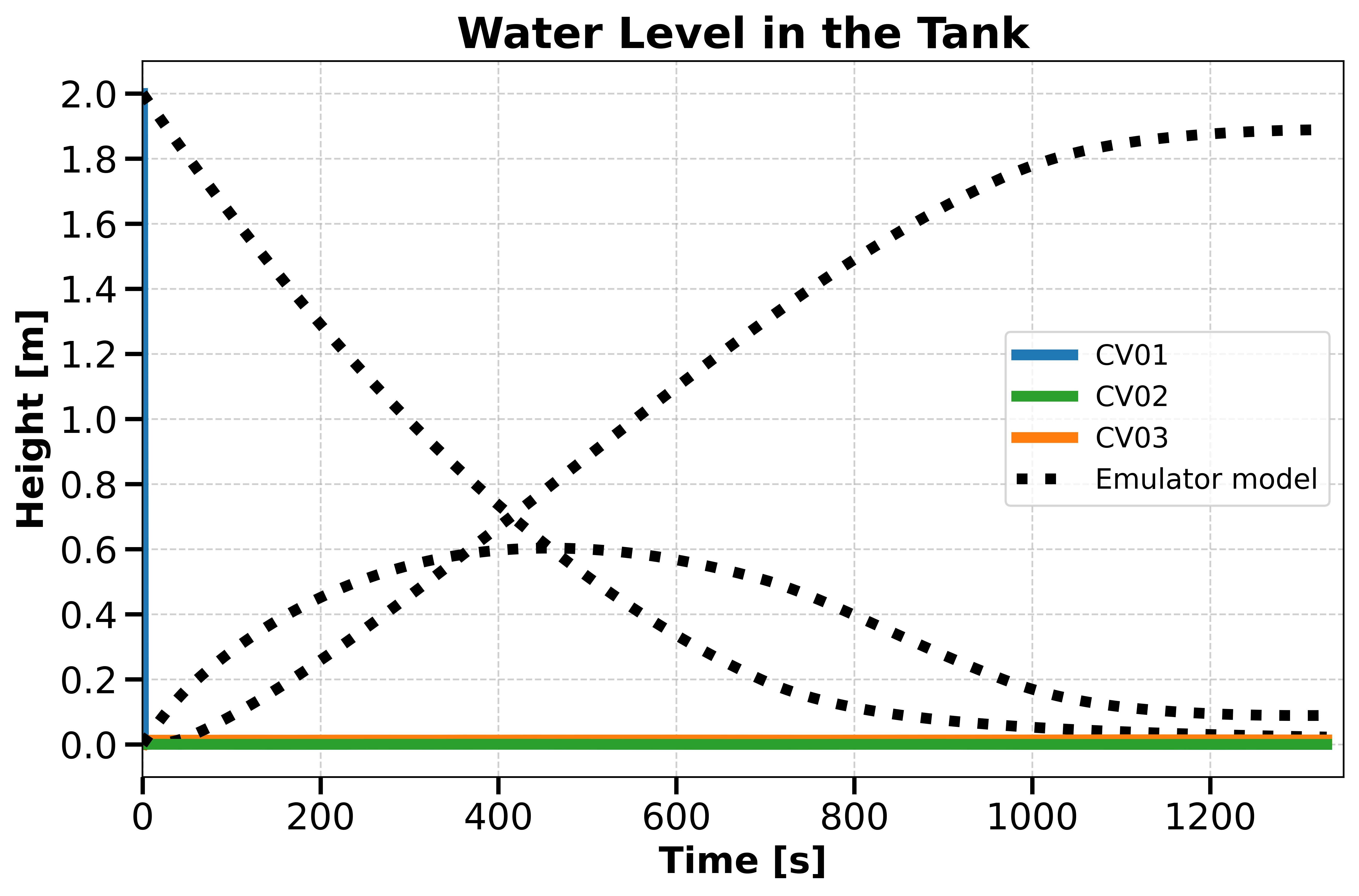}
            \caption{Height profile when $N_i=3$}
        \end{subfigure}
        \hfill
        \begin{subfigure}{0.45\textwidth}
            \centering
            \includegraphics[width=\textwidth]{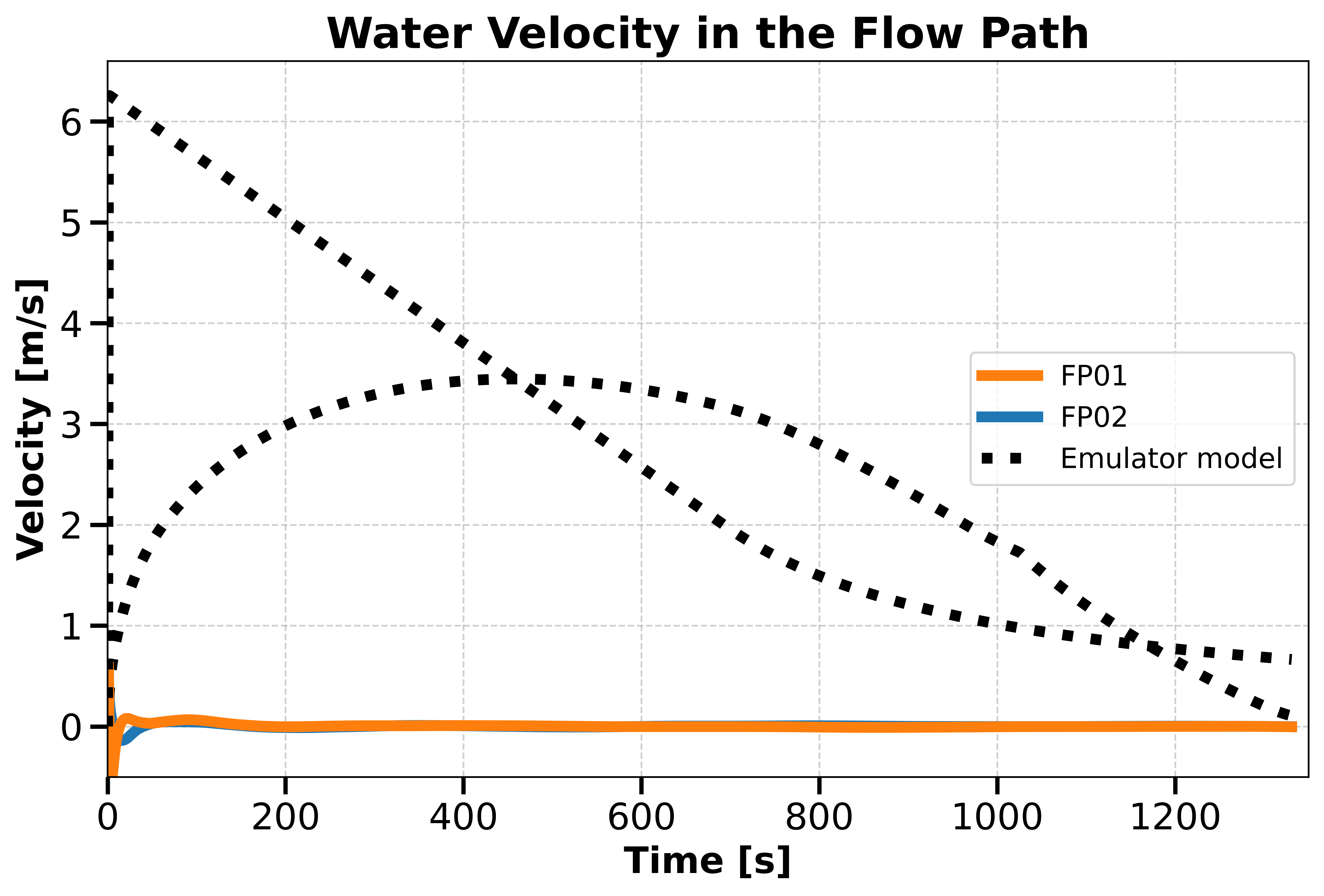}
            \caption{Velocity profile when $N_i=3$}
        \end{subfigure}
    \end{subfigure}
    
    \vspace{0.3cm}
    
    \begin{subfigure}{\textwidth}
        \centering
        \begin{subfigure}{0.45\textwidth}
            \centering
            \includegraphics[width=\textwidth]{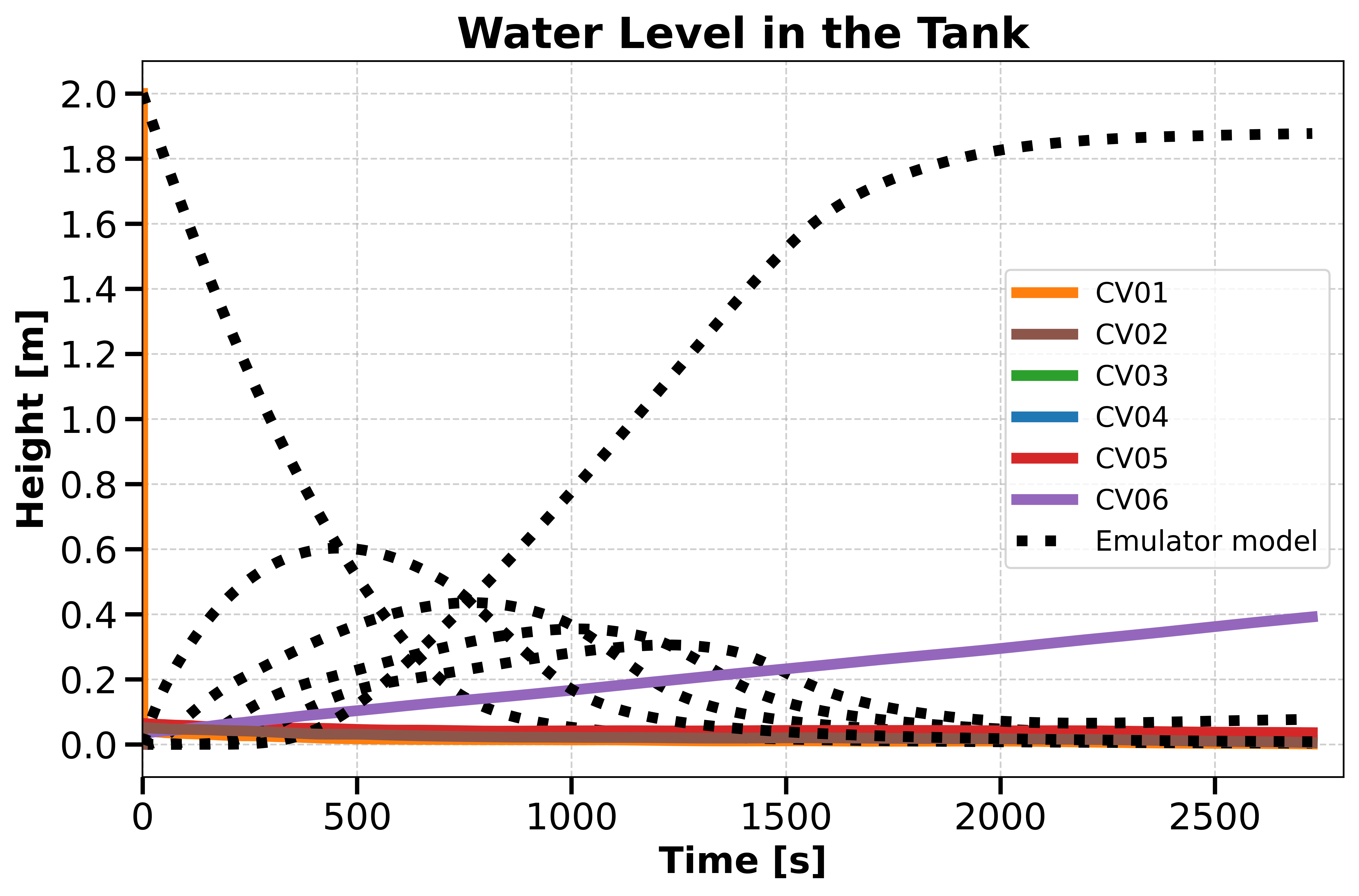}
            \caption{Height profile when $N_i=6$}
        \end{subfigure}
        \hfill
        \begin{subfigure}{0.45\textwidth}
            \centering
            \includegraphics[width=\textwidth]{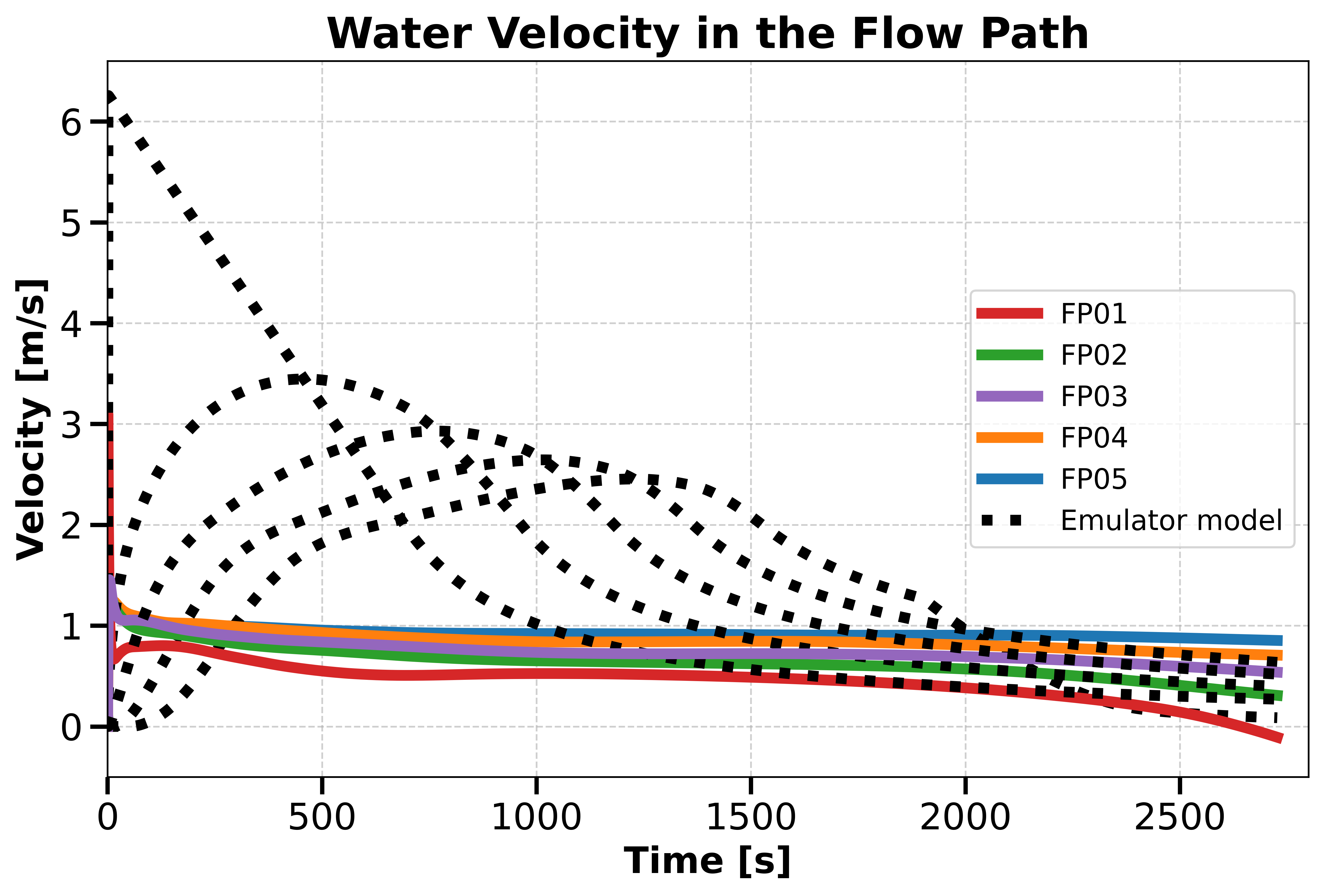}
            \caption{Velocity profile when $N_i=6$}
        \end{subfigure}
    \end{subfigure}
    
    \caption{Comparison of water height (left) and velocity (right) for $N_i = 2$, $3$, and $6$ CV cases. Dotted lines represent the emulator model results, and solid lines indicate the PINN predictions.}
    \label{fig:pinn_comparison}
\end{figure}

Figure~\ref{fig:pinn_comparison} presents the comparison results for the cases with $N_i = 2$, $3$, and $6$ CVs. The figure displays the time histories of water height and velocity for each case, where the Python-based emulator model results are represented by dotted lines and the PINN predictions by solid lines. In all cases, PINN totally fails to capture the transient behavior, exhibiting significant discrepancies compared to the reference solutions. The errors remain substantial regardless of the system complexity, indicating a fundamental limitation in the PINN’s ability to learn the governing dynamics even in the simplest case, $N_i=2$.

The corresponding MAE and MSE values confirm the poor accuracy of the PINN across all cases, as summarized in Table ~\ref{tab:comparison}. A detailed analysis of the PINN results reveals that, in certain cases, the predicted solutions remain nearly constant over time, failing to capture the expected transient behavior of the system. This issue was observed consistently across all tested configurations, where the network does not exhibit meaningful variations despite changing input conditions. The observed behavior suggests that the vanilla PINN struggles to learn the complex interactions between CVs, likely due to its single-network architecture, which must simultaneously predict multiple interdependent variables as in Figure~\ref{fig:pinn_architecture}. As a result, the network fails to establish an appropriate mapping between inputs and outputs, leading to a trivial solution where the predicted state remains unchanged. 

\begin{table}[htbp]
    \centering
    \begin{adjustbox}{max width=\textwidth}
    \begin{tabular}{ccccccc}
        \toprule
        $N_i$ & Architecture & MAE (H) & MSE (H) & MAE (V) & MSE (V) & Parameters \\
        \midrule
        2 & PINN & 2.656295 & 4.870166 & 3.252238 & 13.065767 & 334,467 \\
        2 & NA-PINN & 0.002709 & 6.622294e-06 & 0.008585 & 2.142900e-04 & 337,907 \\
        3 & PINN & 1.985598 & 2.420244 & 5.039016 & 16.950108 & 593,925 \\
        3 & NA-PINN & 0.003791 & 6.184979e-06 & 0.010429 & 2.332637e-04 & 579,845 \\
        6 & PINN & 1.678132 & 1.640074 &  4.425523 & 8.593361 & 1,226,923 \\
        6 & NA-PINN & 0.006960 & 2.452554e-05 & 0.024328 & 3.906309e-04 & 1,275,659 \\
        \bottomrule
    \end{tabular}
    \end{adjustbox}
    \caption{Comparison of MAE, MSE, and the number of trainable parameters between the vanilla PINN and the NA-PINN.}
    \label{tab:comparison}
\end{table}

\begin{figure}[H]
    \centering
    \begin{subfigure}{0.30\textwidth}
        \centering
        \includegraphics[width=\textwidth]{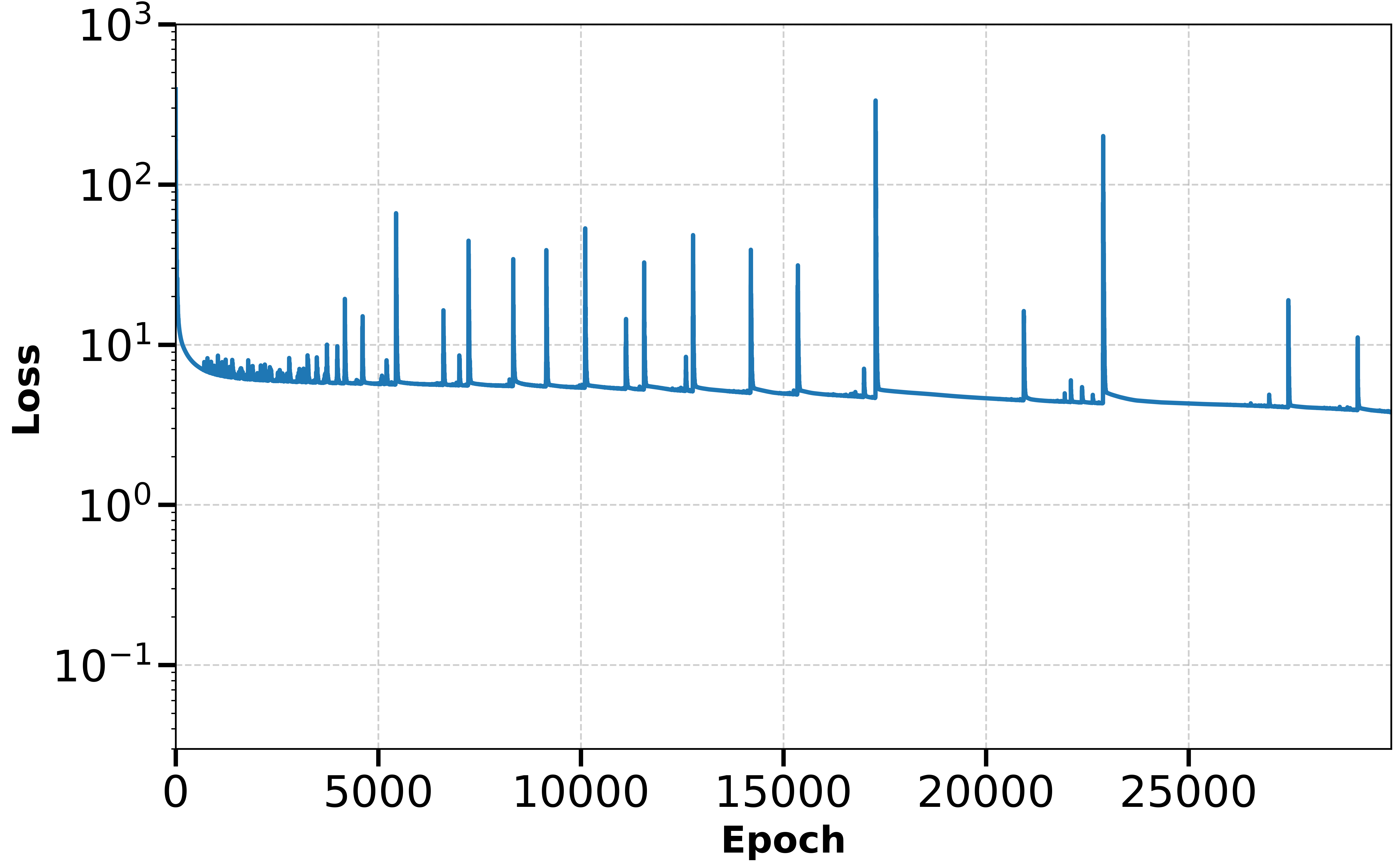}
        \caption{Training loss at $N_i = 2$}
    \end{subfigure}
    \hfill
    \begin{subfigure}{0.30\textwidth}
        \centering
        \includegraphics[width=\textwidth]{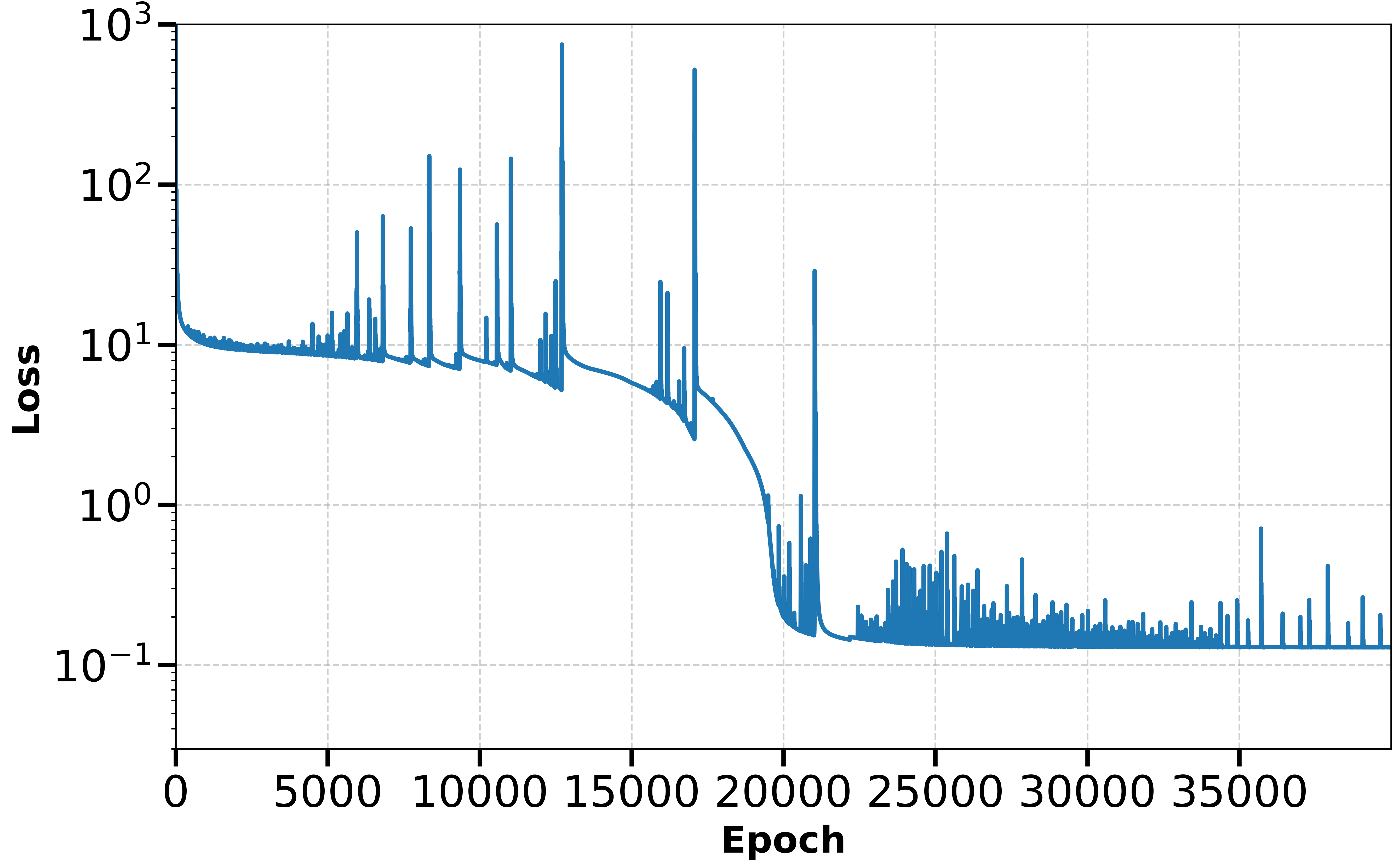}
        \caption{Training loss at $N_i = 3$}
    \end{subfigure}
    \hfill
    \begin{subfigure}{0.30\textwidth}
        \centering
        \includegraphics[width=\textwidth]{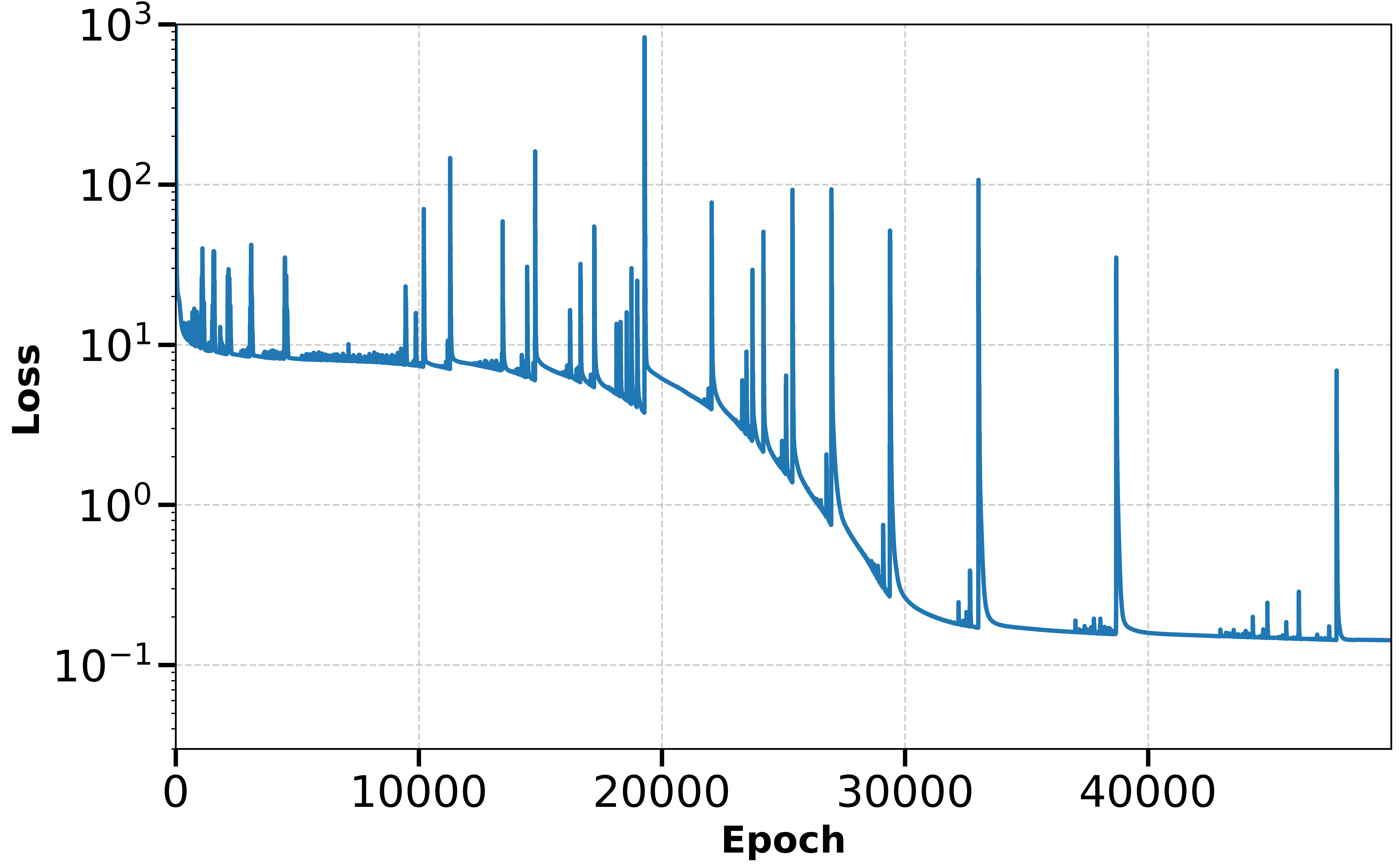}
        \caption{Training loss at $N_i = 6$}
    \end{subfigure}    
    \caption{Training loss histories for $N_i = 2$, $3$, and $6$ CV cases. The total loss is plotted over the training epochs, showing no meaningful convergence in any case.}
    \label{fig:loss_pinn}
\end{figure}

In addition, Figure~\ref{fig:loss_pinn} presents the loss histories during training for each case. For $N_i = 2$, the loss curve shows no noticeable decrease, indicating a clear failure to converge. In contrast, the $N_i = 3$ and $N_i = 6$ cases exhibit a reduction in loss by approximately two orders of magnitude from the initial epochs. However, these cases appear to converge to local minima, as the solutions fail to reflect the expected physical behavior. This limitation is likely due to the increased number of PDEs---five for $N_i = 3$ and eleven for $N_i = 6$---which poses a significant challenge for the vanilla PINN architecture based on a single neural network. These results highlight the difficulty of solving highly coupled multi-equation systems using the vanilla PINN approach.

These results clearly demonstrate that the vanilla PINN formulation struggles to model the transient TH behavior of the system considered in this study. This highlights the necessity for architectural improvements and more advanced training strategies, which will be discussed in the following chapter.

\section{Node-assigned PINN for thermal-hydraulics analysis code}\label{sec:4}

\subsection{Concept and architecture}\label{sec:4.1}

The vanilla PINN architecture exhibits significant limitations when applied to transient TH simulations. These limitations arise because a single neural network is used to predict multiple PDEs simultaneously, which can lead to gradient conflicts and convergence issues due to multiple local minima and differences in the output variable scales~\cite{l-hydra}. Moreover, strong interdependencies between CVs and FLs can result in convergence failure and large prediction errors. We concluded that a novel PINN architecture suitable for CV-based system codes needs to be developed.

\begin{figure}[H]
    \centering
    \includegraphics[width=1.0\textwidth]{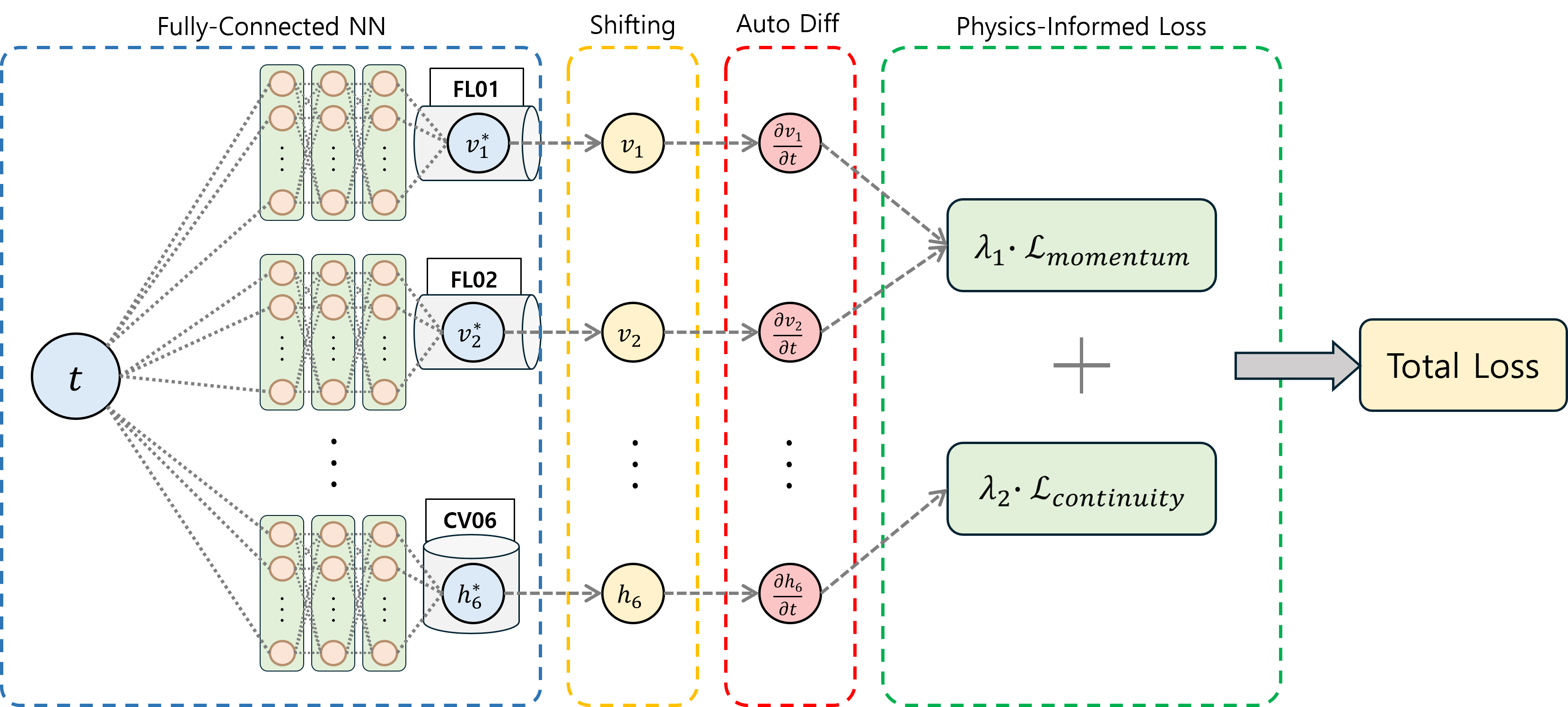}
    \caption{Architecture of the NA-PINN. Each network is responsible for predicting one CV or FL variable, as illustrated inside each block.}
    \label{fig:NBP_architacture}
\end{figure}

To address these challenges, we propose a new architecture referred to as the NA-PINN. In this framework, each output variable, including water height and velocity, is assigned an independent neural network. In other words, separate networks were assigned for modeling individual nodes of CVs and FLs. It is named node, not CV, because it also includes FL objects. Although the networks are structurally independent, they are trained simultaneously via a shared loss function that enforces the governing physical laws, ensuring consistency among variables during training. A schematic of the NA-PINN architecture is shown in Figure~\ref{fig:NBP_architacture}.

In vanilla PINN, only temporal information is used as an input, a single neural network must infer spatial-temproal functions although spatially descrete . By contrast, in the NA‑PINN architecture each subnetwork is associated with a specific spatial node, so that, given only the temporal input, it predicts a purely temporal function. We attribute the superior performance of NA‑PINN to the deliberate exclusion of spatial information from both its inputs and outputs, which simplifies the learning task .

\subsection{Node-assigned PINN module verification}\label{sec:4.2}

The architecture of NA-PINN addresses the limitations associated with using a single network as in Section~\ref{sec:3.4} by assigning a individual neural network to each output variable. This approach eliminates cross-output interference and enables each network to specialize in its respective variable. Moreover, the shared loss function ensures that the interdependence among variables governed by physical laws is effectively captured during training ~\cite{zhang2023cpinns}. Apart from this structural modification, the training settings, including the optimizer, activation function, and learning rate, are consistent with those used in the vanilla PINN.

The performance of the NA-PINN module is validated through numerical experiments conducted on systems with \(N_i = 2, 3,\) and \(6\) CVs---same as vanilla PINN. To ensure a fair comparison, the total number of trainable parameters in NA-PINN is kept similar to that of the vanilla PINN. The hyperparameters applied in each case are summarized in Table~\ref{tab:table_of_hyperparameters_of_NBP}, where the number of networks in each case is set equal to the number of output variables to be predicted—that is, the number of PDEs associated with water heights and velocities.

\begin{table}[h!]
    \centering
    \begin{adjustbox}{max width=\textwidth}
    \begin{tabular}{ccccccc p{3.5cm}}
        \toprule
        \textbf{$N_i$} & \textbf{Networks} & \textbf{End Time} & \textbf{Collocation Points} & \textbf{Epochs} & \textbf{Hidden Layers $\times$ Nodes} \\
        \midrule
        2 & 3 & 1000 & 2500 & 30000 & $8 \times 128$ \\
        3 & 5 & 1400 & 3000 & 40000 & $8 \times 128$ \\
        6 & 11 & 2800 & 6000 & 50000 & $8 \times 128$ \\
        \bottomrule
    \end{tabular}
    \end{adjustbox}
    \caption{Hyperparameters used for the NA-PINN module.}
    \label{tab:table_of_hyperparameters_of_NBP}
\end{table}

\begin{figure}[H]
    \centering
    \begin{subfigure}{\textwidth}
        \centering
        \begin{subfigure}{0.45\textwidth}
            \centering
            \includegraphics[width=\textwidth]{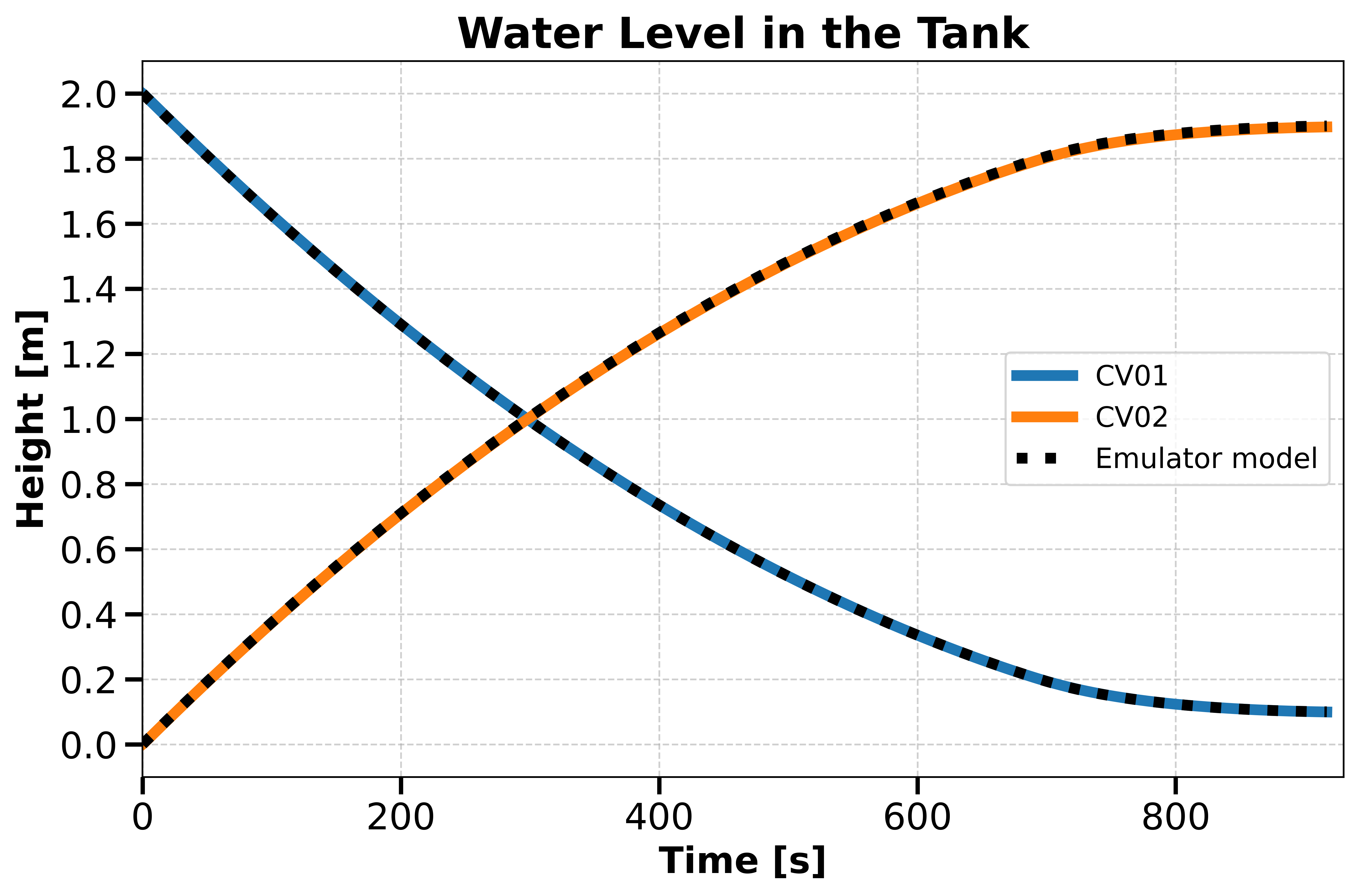}
            \caption{Height profile when $N_i=2$}
        \end{subfigure}
        \hfill
        \begin{subfigure}{0.45\textwidth}
            \centering
            \includegraphics[width=\textwidth]{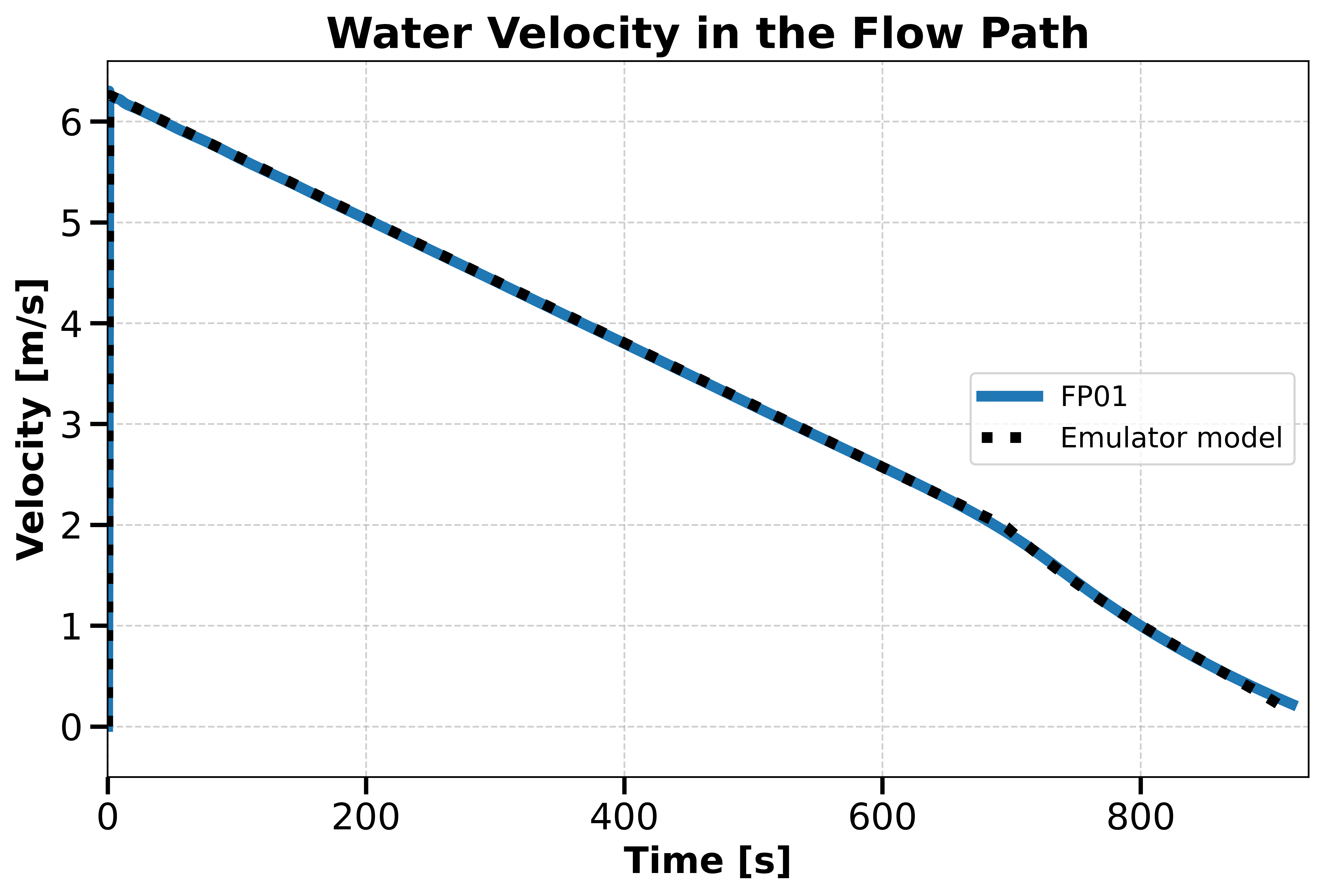}
            \caption{Velocity profile when $N_i=2$}
        \end{subfigure}
    \end{subfigure}
    
    \vspace{0.3cm}
    
    \begin{subfigure}{\textwidth}
        \centering
        \begin{subfigure}{0.45\textwidth}
            \centering
            \includegraphics[width=\textwidth]{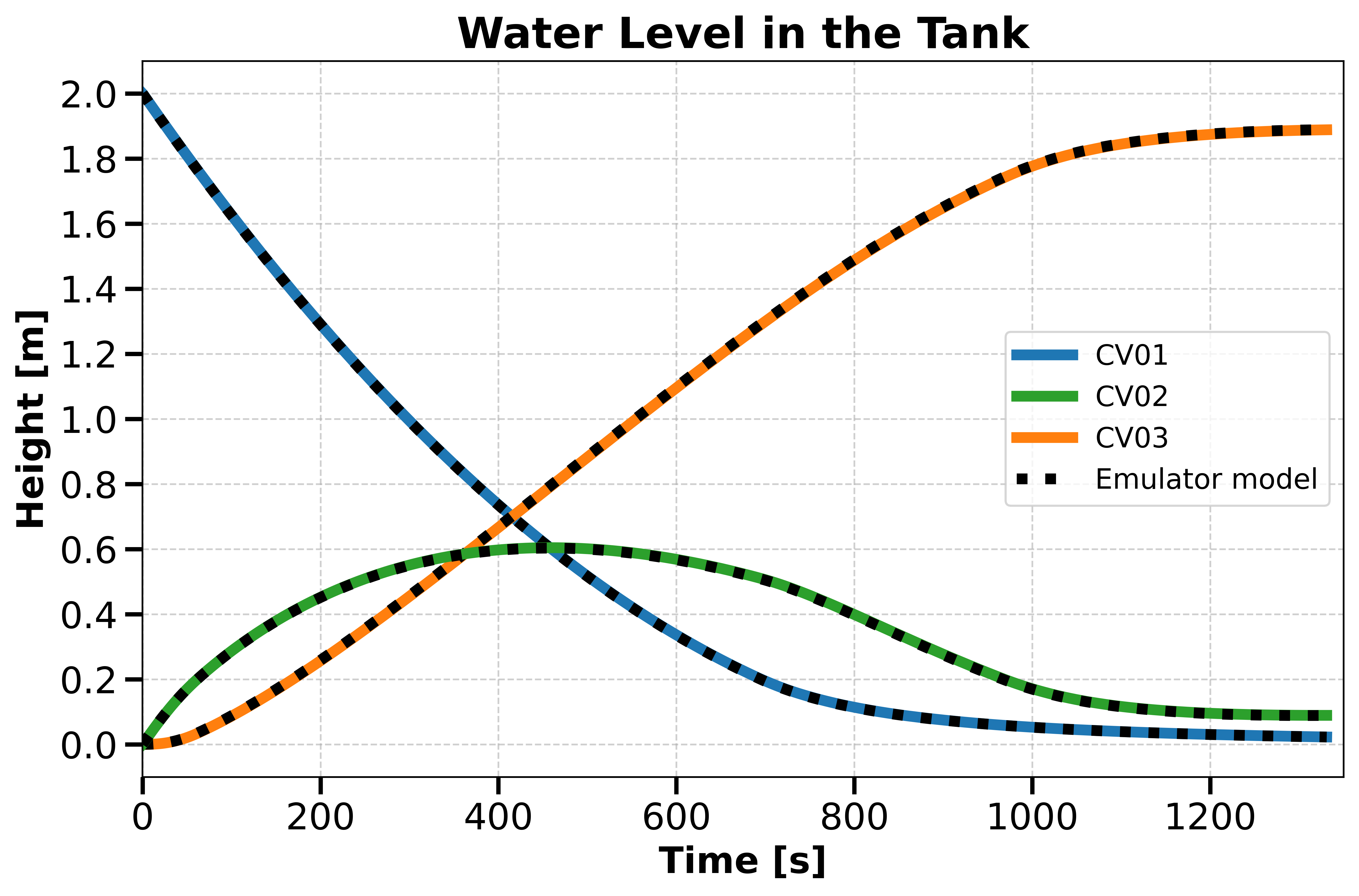}
            \caption{Height profile when $N_i=3$}
        \end{subfigure}
        \hfill
        \begin{subfigure}{0.45\textwidth}
            \centering
            \includegraphics[width=\textwidth]{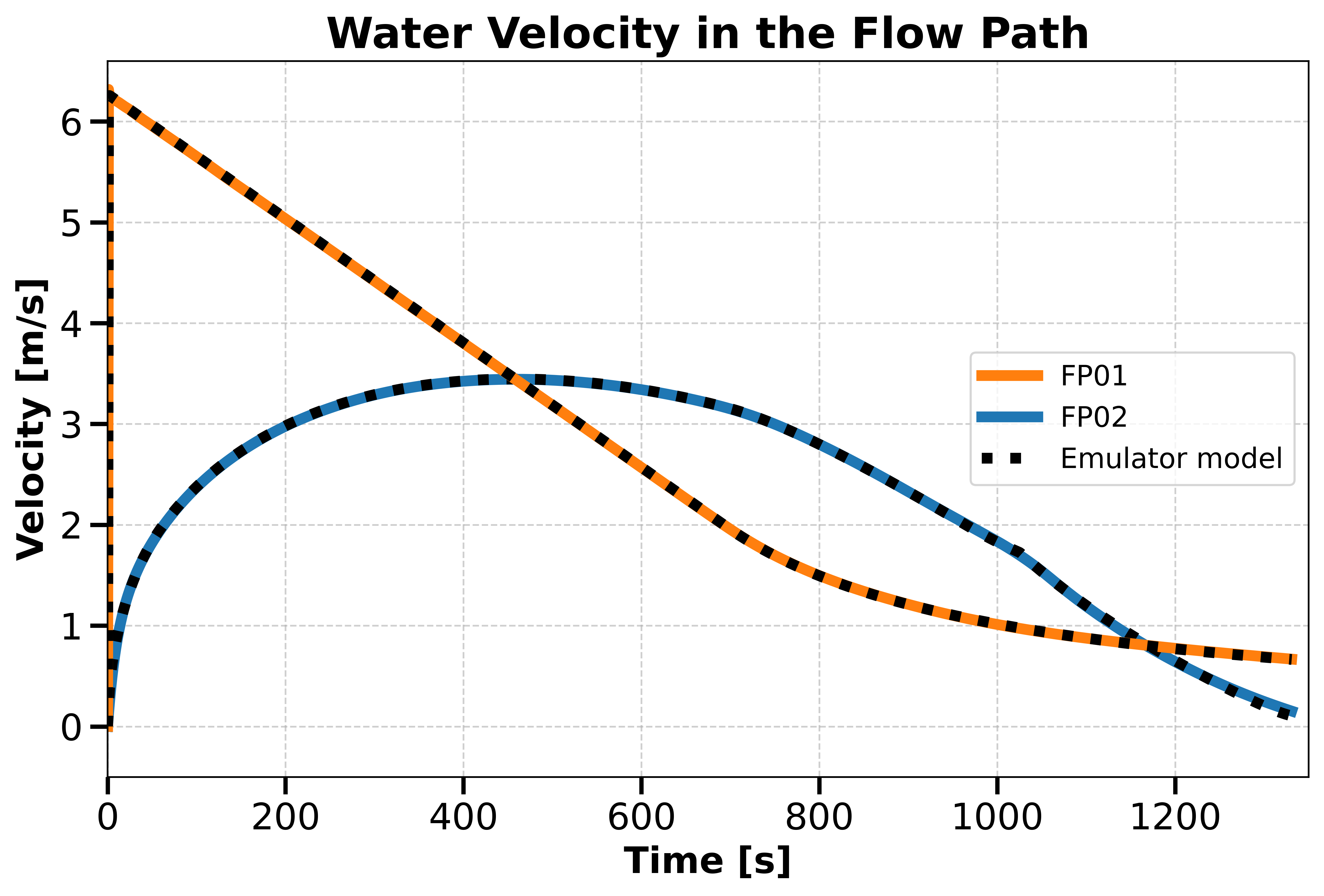}
            \caption{Velocity profile when $N_i=3$}
        \end{subfigure}
    \end{subfigure}
    
    \vspace{0.3cm}
    
    \begin{subfigure}{\textwidth}
        \centering
        \begin{subfigure}{0.45\textwidth}
            \centering
            \includegraphics[width=\textwidth]{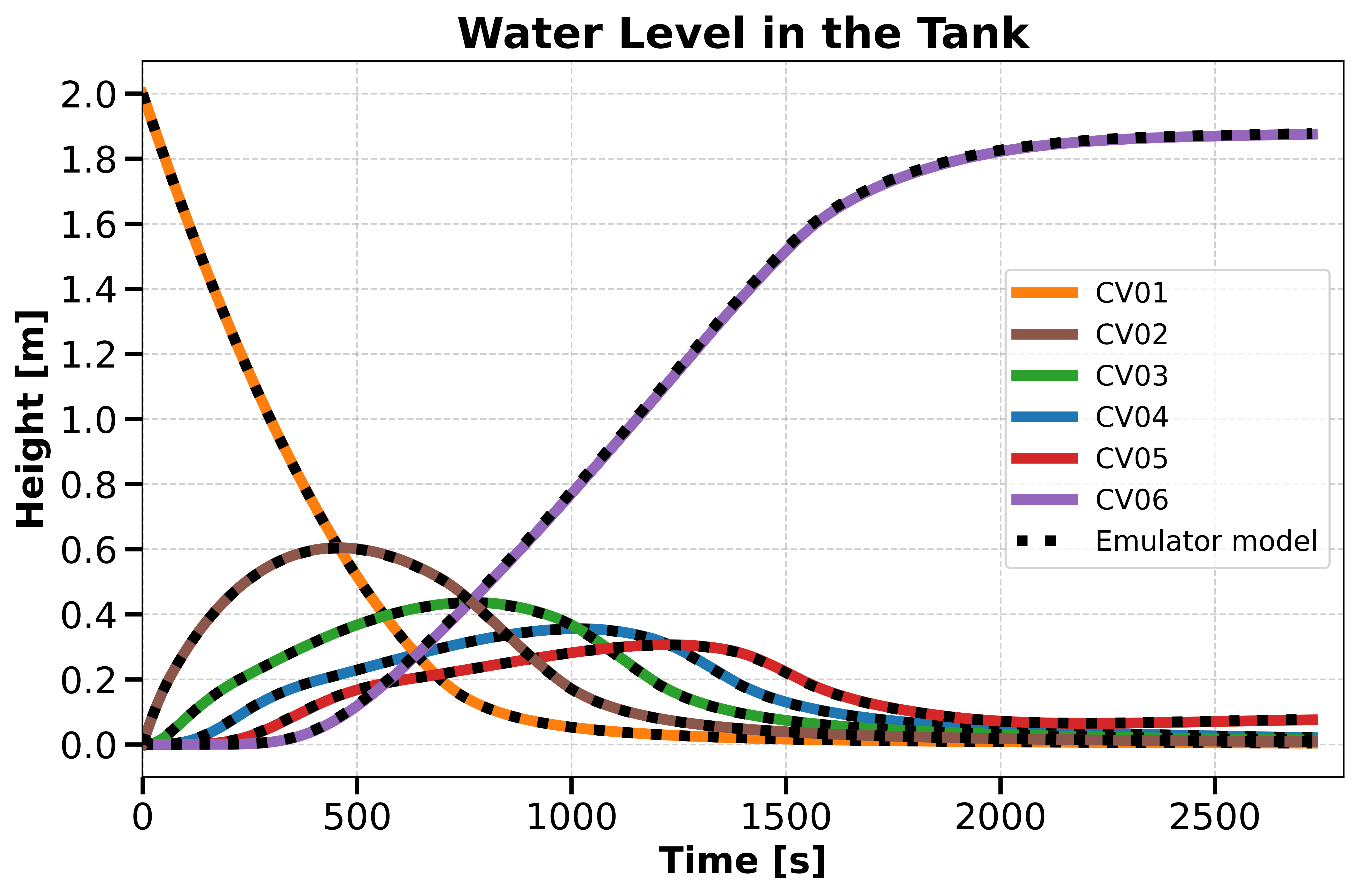}
            \caption{Height profile when $N_i=6$}
        \end{subfigure}
        \hfill
        \begin{subfigure}{0.45\textwidth}
            \centering
            \includegraphics[width=\textwidth]{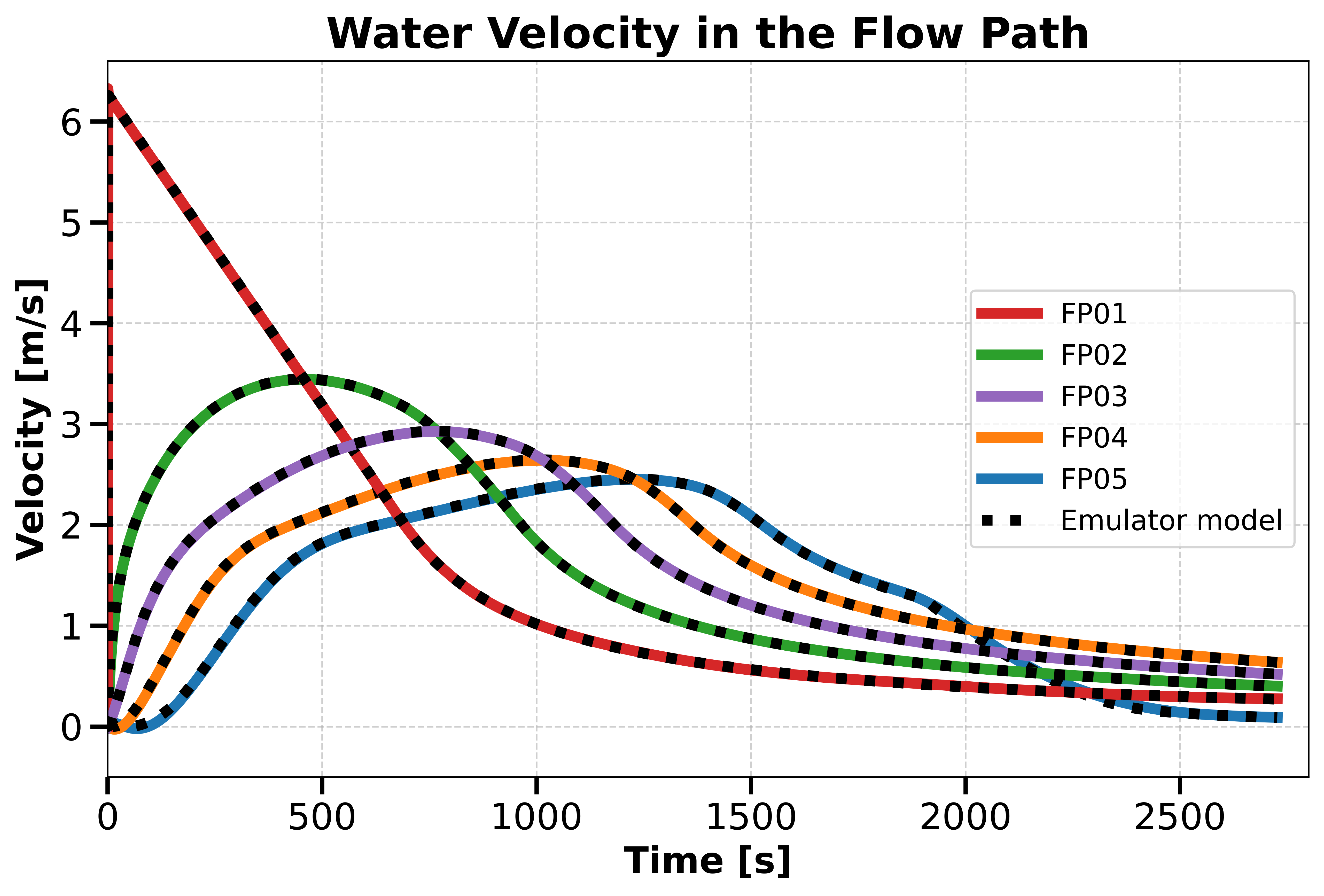}
            \caption{Velocity profile when $N_i=6$}
        \end{subfigure}
    \end{subfigure}
    
    \caption{Comparison of water height (left) and velocity (right) for $N_i = 2$, $3$, and $6$ CV cases using the NA-PINN. Dotted lines represent the Python-based emulator model results, and solid lines indicate the NA-PINN predictions.}
    \label{fig:NBP_comparison}
\end{figure}

Figure~\ref{fig:NBP_comparison} presents their results, showing that NA-PINN closely follows the reference solutions obtained from the Python-based emulator model. In contrast to the vanilla PINN, which struggled with learning meaningful dynamics, NA-PINN accurately captures transient variations. This improvement can be attributed to its decoupled network architecture, which eliminates interference between outputs and allows each neural network to specialize in learning its assigned variable.

Table~\ref{tab:comparison} quantitatively confirms that NA-PINN significantly reduces MAE and MSE across all test cases. Unlike the vanilla PINN, which failed to learn transient behavior, NA-PINN effectively models dynamic variations by leveraging its independent network structure. This structural improvement enables more stable gradient updates, leading to more accurate predictions in complex simulations.

\begin{figure}[H]
    \centering
    \begin{subfigure}{0.30\textwidth}
        \centering
        \includegraphics[width=\textwidth]{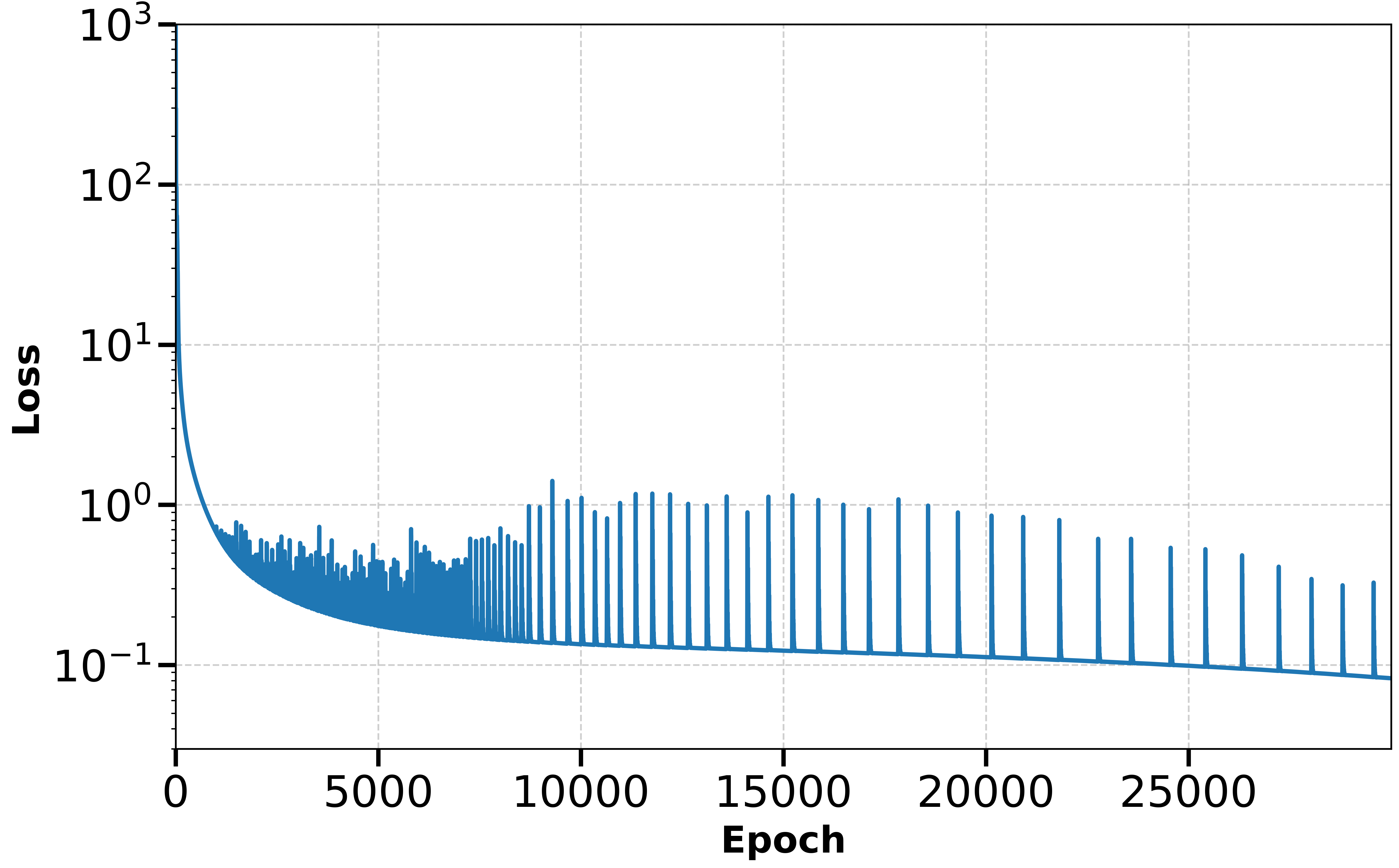}
        \caption{Training loss at $N_i = 2$}
        \label{fig:example1}
    \end{subfigure}
    \hfill
    \begin{subfigure}{0.30\textwidth}
        \centering
        \includegraphics[width=\textwidth]{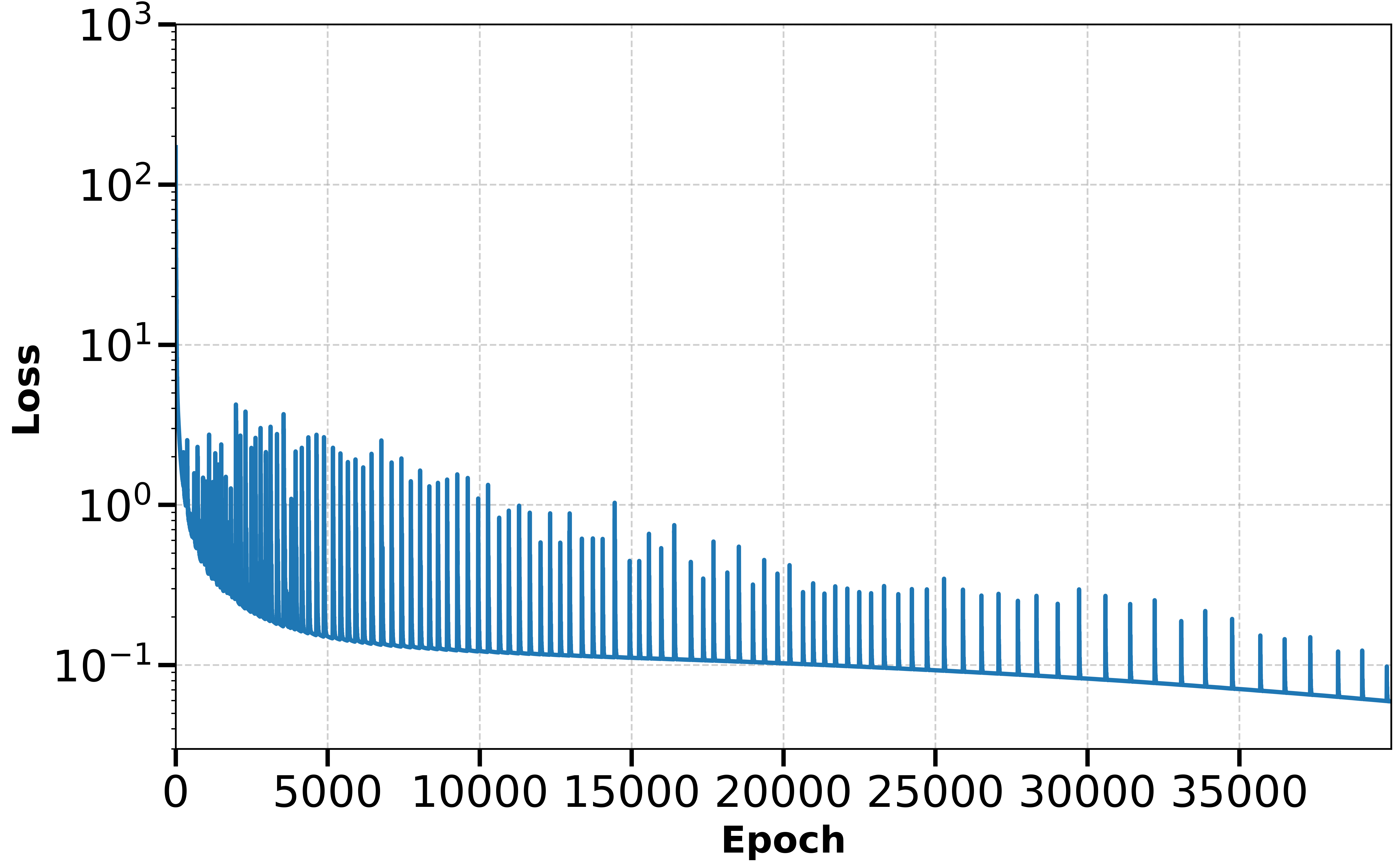}
        \caption{Training loss at $N_i = 3$}
        \label{fig:example2}
    \end{subfigure}
    \hfill
    \begin{subfigure}{0.30\textwidth}
        \centering
        \includegraphics[width=\textwidth]{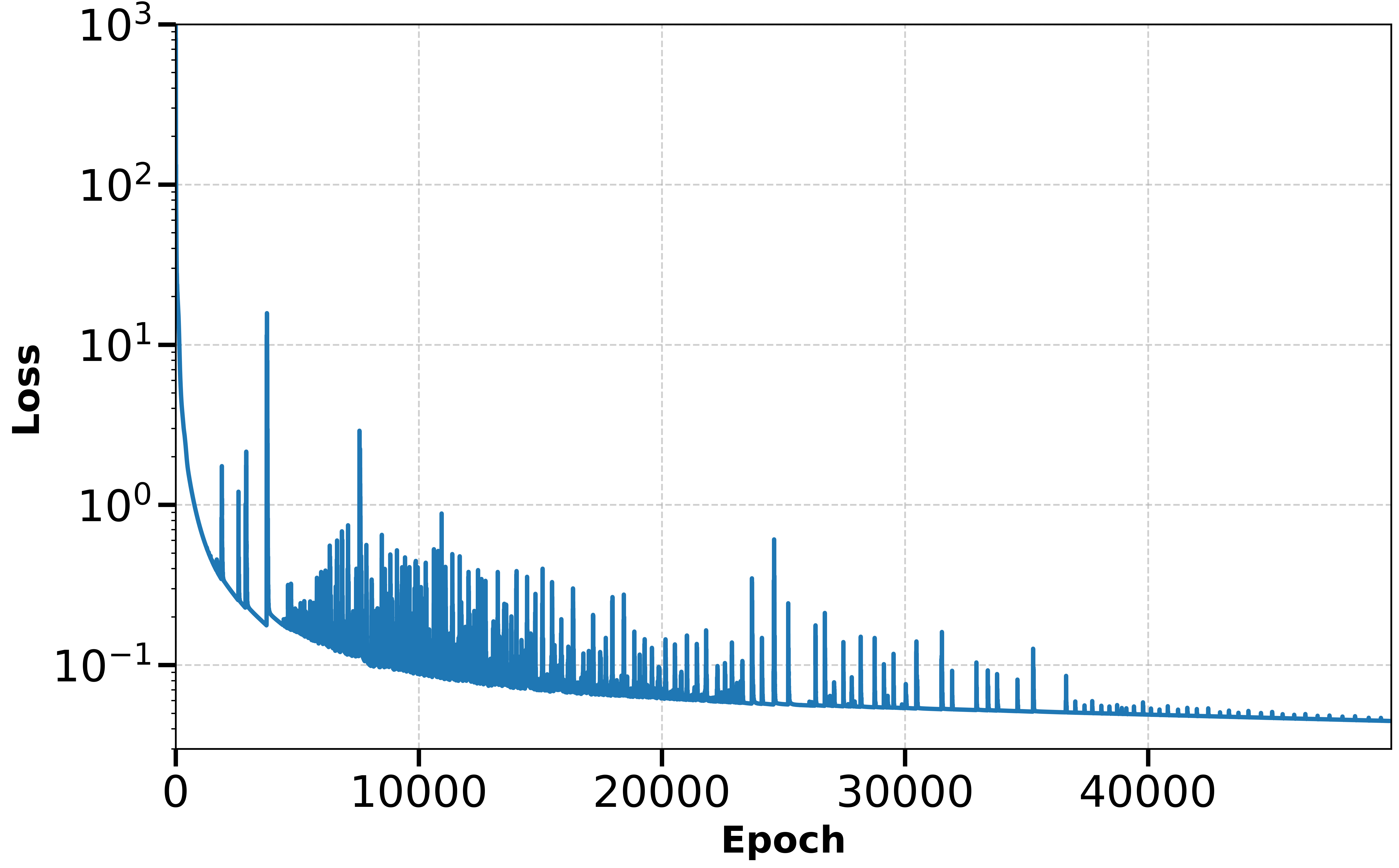}
        \caption{Training loss at $N_i = 6$}
        \label{fig:example2}
    \end{subfigure}    
    \caption{Training loss histories of the NA-PINN module for $N_i = 2$, $3$, and $6$ CV cases.}
    \label{fig:loss_NBP}
\end{figure}

Additionally, Figure~\ref{fig:loss_NBP} presents the training loss histories for each case, further demonstrating the advantages of NA-PINN. All cases exhibit a stable and gradual decrease in training loss, suggesting that the model progressively converges toward a global minimum. This contrasts with the stagnation observed in Figure~\ref{fig:loss_pinn}, where the loss fails to improve due to convergence to local minimum. These results highlight the effectiveness of the proposed NA-PINN architecture, which facilitates better optimization by avoiding poor local minimum and guiding the training process toward more accurate solutions.

\section{Future Work}\label{sec:5}
While the proposed NA-PINN architecture demonstrates improved performance over the vanilla PINN, several directions remain for future investigation.
First, the scenarios used in this study assume a simplified flow pattern along CVs and FLs. Therefore, further validation is needed for more complex and realistic transient scenarios.
Second, the current implementation focuses solely on MELCOR's CVH/FL package. Future work could involve extending the framework to include the Heat Structure (HS) package or coupling with other MELCOR modules, such as the fission product module, to develop a more comprehensive multi-physics solver.
Third, the current model requires retraining whenever the scenario changes, which limits its applicability in real-time or fast-response simulations. This limitation could be addressed by developing a surrogate PINN framework capable of generalizing across a range of accident scenarios after a single training process.

\section{CONCLUSION}\label{sec:6}
Severe accident analysis code enables to investigate the progression of the accident by modeling the postulated scenarios. One of the most representative code, MELCOR is obtained as a target analysis system code in this study. MELCOR has a wide range of application from solving numerical solutions for TH and heat structure to modeling complex severe accident progression. However, MELCOR still faces a few challenges such as manually constructing the complex inputs and being unable to solve the multiphysics problem. The goal of this study is to address these system codes difficulties by employing artificial intelligence. As a pioneering study, the CVH/FL package, a simple yet crucial component, was simulated using a Python-based emulator. to thoroughly assess the comprehension of the logic behind the calculation in the package. The comparison results indicated that the system code and the emulator well agreed between the two. With the emulator being developed, reduction in complexity of the loss term for PINN model could be conducted thereby finding the optimized loss equation for PINN model to train. 

We proposed a NA-PINN architecture to address the limitations of the vanilla PINN in modeling transient TH systems. By assigning independent neural networks to each PDE and enforcing a shared loss function, the proposed method effectively captures the coupled dynamics of CVs and FLs while removing spatial information from its inputs and outputs, which simplifies the learning task into a purely temporal approximation and results in superior performance. Numerical experiments demonstrated that NA-PINN achieved significantly better convergence and accuracy compared to vanilla PINN, particularly in systems with multiple interacting PDEs. At $N_i = 2$, the MSE for height decreased from 4.87 to $6.62 \times 10^{-6}$, and that for velocity decreased from 13.07 to $2.14 \times 10^{-4}$. At $N_i = 3$, the MSE for height decreased from 2.40 to $6.18 \times 10^{-6}$, while the velocity MSE decreased from 16.95 to $2.33 \times 10^{-4}$. When $N_i = 6$, the MSE for height decreased from 1.64 to $2.45 \times 10^{-5}$, and for velocity from 8.59 to $3.91 \times 10^{-4}$. These results demonstrate that NA-PINN achieved substantially lower errors than vanilla PINN for both height and velocity across all tested scenarios.

The gradual loss reduction and physically meaningful predictions highlight the advantages of the architecture.
These results highlight the potential of NA-PINN as a scalable and reliable approach that could be extended to complex multi-physics problems in nuclear THs.

\section*{Acknowledgments}
This work was supported by the Ministry of Science and ICT of Korea (No. RS-2024-00355857) and by the Nuclear Safety Research Program through the Korea Foundation of Nuclear Safety (KoFONS), funded by the Nuclear Safety and Security Commission (NSSC) of the Republic of Korea (No. RS-2024-00403364).


\bibliographystyle{elsarticle-num} 
\bibliography{sample}






\end{document}